\documentclass[10pt,twocolumn,letterpaper]{article}

\usepackage{iccv}              % To produce the CAMERA-READY version
\definecolor{iccvblue}{rgb}{0.21,0.49,0.74}
\usepackage[pagebackref,breaklinks,colorlinks,allcolors=iccvblue]{hyperref}

\usepackage{etoc}
\usepackage{etoc} % Customization of the table of contents
\usepackage{titletoc}
\usepackage[subfigure]{tocloft} %模板中用了subfigure，不加此选项会产生冲突

\usepackage{epsfig}
\usepackage{graphicx}
\usepackage{amsmath}
\usepackage{amssymb}
\usepackage{booktabs}
\usepackage{multirow}
\usepackage{tabularx}
\usepackage{enumitem}
\usepackage{gensymb}
\usepackage{colortbl}
\usepackage{caption}
\usepackage{makecell}
\usepackage{wrapfig}
\usepackage{amssymb}% http://ctan.org/pkg/amssymb
\usepackage{pifont}% http://ctan.org/pkg/pifont
\definecolor{mycyan}{cmyk}{.1,0,0,0}
\newcommand{\mypara}[1]{\vspace{1mm}\noindent\textbf{#1}}
\usepackage[capitalize]{cleveref}
\newcommand{\name}{RoboPearls}

\newcommand{\tr}[2][\@empty]{%
  \ifx#1\@empty
    \textcolor{red}{#2}%
  \else
    \textcolor{red}{\fontsize{#1}{#1}\selectfont #2}%
  \fi
}
\newcommand{\ts}[2][\@empty]{%
  \ifx#1\@empty
    {#2}%
  \else
    {\fontsize{#1}{#1}\selectfont #2}%
  \fi
}
\usepackage{colortbl}

\definecolor{mycyan}{cmyk}{.1,0,0,0}
\definecolor{mygray}{gray}{.95}
\definecolor{mypink}{rgb}{.99,.91,.95}
\definecolor{mypurple}{rgb}{0,0,0}

\definecolor{F7E0D5}{RGB}{245,240,255}

\colorlet{Light}{White!0!F7E0D5}

\usepackage{adjustbox}
\definecolor{ImportantColor}{rgb}{0.63, 0.79, 0.95}

\def\paperID{2317} % *** Enter the Paper ID here
\def\confName{ICCV}
\def\confYear{2025}

\title{\raisebox{-1.6ex}{\includegraphics[height=2.0\baselineskip]{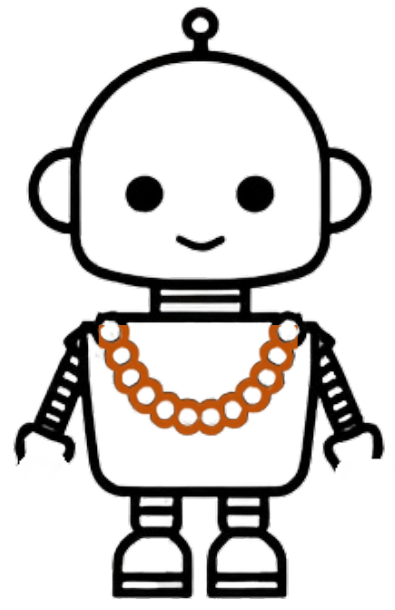}}  \name{}: Editable Video Simulation for Robot Manipulation}

\author{
  Tang Tao$^{1}$$^{*\ddagger}$ \quad
  Likui Zhang$^{2}$$^*$ \quad
  Youpeng Wen$^{1}$ \quad
  Kaidong Zhang$^{2}$ \quad 
  Jia-Wang Bian$^{3}$ \quad 
  Xia Zhou$^{4}$ \quad \\
  Tianyi Yan$^{4}$ \quad 
  Kun Zhan$^{4}$ \quad
  Peng Jia$^{4}$ \quad
  Hefeng Wu$^{2}$ \quad
  Liang Lin$^{2}$ \footnotemark[2]\quad
  Xiaodan Liang$^{1}$ \footnotemark[2] \quad
  \\
\fontsize{11pt}{\baselineskip}\selectfont
$^1$ Shenzhen Campus of Sun Yat-sen University
\fontsize{11pt}{\baselineskip}\selectfont
$^2$  Sun Yat-sen University \\
\fontsize{11pt}{\baselineskip}\selectfont
$^3$ Bytedance Seed
\fontsize{11pt}{\baselineskip}\selectfont
$^4$ Li Auto Inc. \\
 \begin{normalsize}${\tt trent.tangtao@gmail.com, zhanglk9@mail2.sysu.edu.cn} $\end{normalsize}
}

\begin{document}
% \maketitle

% \begin{figure*}[t]
%     \centering
%     \includegraphics[width=1\linewidth]{imgs/teaser.pdf}
%     \caption{\textbf{The misalignment issue in multimodal implicit field.}
%     }
%     \label{fig:teaser}
% \end{figure*}

\twocolumn[{ 
\maketitle
  \centering
  \includegraphics[width=1.0\textwidth]{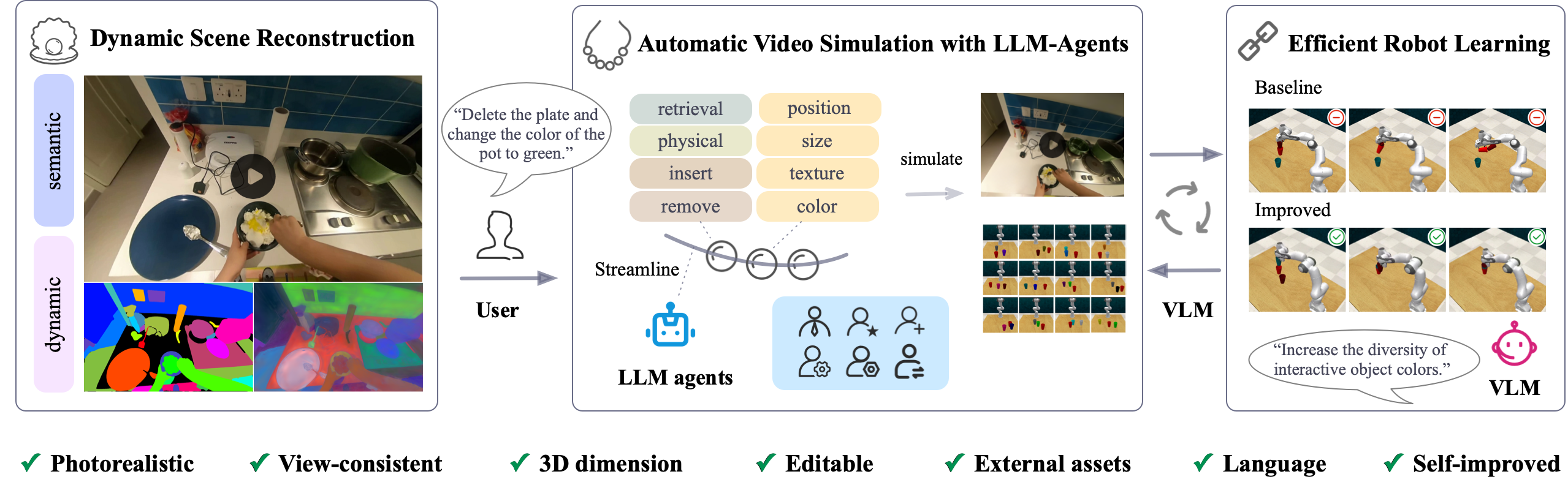}
  \vspace{-6mm}
  \captionof{figure}{\textbf{\name{}, an editable video simulation framework for robotic manipulation.} 
  \name{} reconstructs photo-realistic scenes with semantic features from demonstration videos.
  Then, with various simulation operators, \name{} leverages multiple LLM agents to process user commands into specific editing functions.
  Furthermore, \name{} utilizes a VLM to analyze learning issues and generate corresponding simulation demands to enhance robotic performance.
  % Furthermore, \name{} enhances robotic performance by utilizing VLMs to analyze learning issues and generate corresponding simulation demands to close the simulation loop.
   }
\label{fig:fig1}
\vspace{3mm}
}]

% 图有点空
% 中间的图表达的不是很清楚

\renewcommand{\thefootnote}{}
\footnotetext{$^*$ Equal contribution}
\footnotetext{$^{\ddagger}$Work done during an internship at Li Auto Inc.}
\footnotetext{$^{\dagger}$ Co-corresponding author}

\begin{abstract}
%%%% 突然想到一个非常好的故事
% logo：机器人脖子上套珍珠项链
% 3dgs 扇贝； operator 珍珠； agent 链子；vlm 卡扣
% 其他方法只是在偶尔捡到一个个珍珠，而我们是打磨好珍珠，然后串起来了

% teaser： Photorealistic；3D dim；multi-camera；Editable；External assets；Language；Open-source

The development of generalist robot manipulation policies has seen significant progress, driven by large-scale demonstration data across diverse environments. 
However, the high cost and inefficiency of collecting real-world demonstrations hinder the scalability of data acquisition. 
While existing simulation platforms enable controlled environments for robotic learning, the challenge of bridging the sim-to-real gap remains.
%, compounded by difficulties in scenario-specific data collection and manipulation. 
To address these challenges, we propose \name{}, an editable video simulation framework for robotic manipulation. Built on 3D Gaussian Splatting (3DGS), \name{} enables the construction of photo-realistic, view-consistent simulations from demonstration videos, and supports a wide range of simulation operators, including various object manipulations, powered by proposed modules like Incremental Semantic Distillation (ISD) and 3D regularized NNFM Loss (3D-NNFM). 
Moreover, by incorporating large language models (LLMs), \name{} automates the simulation production process in a user-friendly manner through flexible command interpretation and execution. 
Furthermore, \name{} employs a vision-language model (VLM) to analyze robotic learning issues to close the simulation loop for performance enhancement. 
% To this end, we demonstrate the effectiveness of \name{} through extensive experiments on multiple datasets, including RLBench, Ego4D, and Open X-Embodiment, showing significant improvements in robotic manipulation tasks. 
To demonstrate the effectiveness of \name{}, we conduct extensive experiments on multiple datasets and scenes, including RLBench, COLOSSEUM, Ego4D, Open X-Embodiment, and a real-world robot, which 
% show significant improvements in robotic manipulation tasks.
demonstrate our satisfactory simulation performance.
More information can be found on our
\href{https://tangtaogo.github.io/RoboPearls/}{Project Page}.
% Our contributions provide a systematic pipeline for creating adaptable and scalable simulation environments that enhance robot learning and manipulation.
% Code release: \href{https://github.com/tangtaogo/alignmif}{https://github.com/tangtaogo/alignmif}.

\end{abstract}
\section{Introduction}
\label{sec:intro}

% Remarkable progress has been made in recent years towards
% building generalist real-world robot manipulation policies [6, 50], i.e., policies that can perform a wide range of tasks across many environments and even robot embodiments. These advances are underpinned by large-scale datasets [11, 66] and expressive models [1, 6, 29]. However, evaluating these policies in a scalable and reproducible way remains challenging as 
% Current robot training pipelines rely on humans to provide kinesthetic demonstrations or to program simulation environments. 
% Efficient acquisition of real-world embodied data has been increasingly critical. 
% real-world evaluation is expensive and inefficient. 
% Such human involvement is an important bottleneck towards scaling up robot learning across diverse tasks and environments. We propose

% The field of robotics has witnessed rapid development towards generalist robot manipulation policies~\cite{}.
% Generalist robot manipulators require large-scale demonstrations across diverse environments to learn a wide variety of manipulation skills.
% However, large-scale real-world demonstrations captured by human experts~\cite{} tend to take extremely high costs and fail to scale up the data size in an efficient manner. 
% Though several simulation platforms~\cite{}, e.g., CALVIN, provide a controlled yet versatile environment for developing advanced robot learning methods, the sim-to-real gap remains as a persistent obstacle.

The field of robotics has witnessed rapid advancements in the development of generalist robot manipulation policies~\cite{brohan2022rt, team2024octo, kim2024openvla, goyal2023rvt, liu2024rdt}. These policies rely on large-scale demonstrations across diverse environments to enable robots to acquire a wide range of manipulation skills.
However, collecting large-scale real-world demonstrations performed by human experts~\cite{o2023open, khazatsky2024droid, grauman2022ego4d} is prohibitively expensive and inefficient, making it challenging to scale up the data size. 
While several simulation platforms~\cite{mees2022calvin, mandlekar2021matters, gu2023maniskill2, liu2024libero}, provide controlled and versatile environments for developing advanced robot learning methods, the sim-to-real gap remains a significant obstacle.
Furthermore, collecting or reproducing data for specific scenarios poses considerable challenges. For example, if one wants to replace a cup with a different color in a particular scene, even in simulation, it requires reprogramming the simulation environment.

% Specifically, 3DGS models the environment using a set of 3D Gaussians with learnable parameters, offering an explicit and flexible representation of scenes.
% This flexibility allows for the individual editing of each Gaussian, making it particularly suitable for scene simulations.

On the other hand, the emergence of 3D Gaussian Splatting (3DGS)~\cite{kerbl3Dgaussians} has shown impressive reconstruction quality with high training and rendering efficiency. 
Specifically, 3DGS models the environment using a set of 3D Gaussians with learnable parameters, offering an explicit and flexible representation of scenes.
This flexibility also allows for the individual editing of each Gaussian.
Overall, the explicit representation, high-quality reconstruction, and real-time rendering capabilities of 3DGS have opened up new possibilities to construct photo-realistic simulations from demonstration videos for robotic policy learning.
Although many works~\cite{wu2024recent, chen2024survey,ji2024segment4dgaussians, ye2024gaussian} have explored the reconstruction and editing capabilities of 3DGS and presented some toy demos, developing a systematic pipeline for integrating these techniques to the robotics domain, specifically to expand the capabilities of robotic simulation, remains an unexplored area. 
% and enhance robotic performance
It's like previous works have identified individual operators (``pearls") but has yet to polish them into a comprehensive solution (``creating a sparkling pearl necklace") for robots.

In this paper, we introduce \name{}, an editable video simulation framework, as illustrated in \cref{fig:fig1}, which serves as a fully assembled ``pearl necklace" for robotic manipulation. Specifically, the framework is built upon 3DGS, ensuring the ability to construct photo-realistic and view-consistent simulations from demonstration videos.
To enhance the Gaussian representation (like ``oyster to produce pearls") to accommodate a wide range of simulation operators (``pearls"), we extend it to incorporate temporal propagation, capturing the spatiotemporal dynamics of the scene, and embed semantic features distilled from SAM~\cite{kirillov2023segment} to enable scene understanding capabilities.
Secondly, \name{} refines and polishes various simulation operators (``pearls") to cover diverse everyday scenarios. For example, it enables changes to object color or texture, object removal, and object insertion using external digital assets. Moreover, it also supports physical simulations. 
These functions are powered by carefully designed, non-trivial modules, including the Incremental Semantic Distillation (ISD) module and the 3D regularized NNFM Loss (3D-NNFM).
Thirdly, in contrast to traditional simulations that rely heavily on extensive human intervention, 
% \name{} is designed to be user-friendly and capable of handling complex demands.
users can interact with \name{} using simple natural language commands to generate desired simulations.
Specifically, to string the individual pearls together, we leverage multiple tailored large language model (LLM) agents as the ``string" to automate and streamline the simulation production process. The LLM agents decompose user simulation demands into simplified and concrete commands for specific editing functions.
Finally, to complete the pearl necklace, we integrate a vision-language model (VLM) as the ``clasp" to close the simulation loop, which identifies and analyzes issues in robotic learning and generates corresponding simulation demands to enhance robotic performance.

Through comprehensive evaluations conducted on multiple datasets and scenes, we validate the effectiveness of \name{} in simulating diverse scenarios and
enabling more accurate and robust robotic manipulations.
% across various tasks.
% Specifically, \name{} shows significant improvements on the Colosseum benchmark~\cite{pumacay2024colosseum} (e.g., \texttt{+}20.1 average success scores across all perturbations), which assesses robotics’ generalization across various environmental perturbations. 
% Meanwhile, \name{} achieves state-of-the-art results on the RLBench benchmark~\cite{james2020rlbench} (e.g., \texttt{+}20.1 and \texttt{+}30.1 success scores on the Stack Cups and Push Buttons tasks, respectively).
% Additionally, \name{} consistently performs well on the real-world robot and datasets, including Ego4D~\cite{grauman2022ego4d} and the Open X-Embodiment dataset~\cite{o2023open}.
Specifically, \name{} demonstrates significant improvements on the Colosseum benchmark~\cite{pumacay2024colosseum}, achieving an average success score increase of +17.5 across all perturbations, which highlights its ability to generalize across diverse environmental conditions. 
Furthermore, \name{} achieves state-of-the-art performances on the RLBench benchmark~\cite{james2020rlbench}, with success score gains of +16.4 and +23.0 on the Stack Cups and Put in Cupboard tasks, respectively. 
Additionally, \name{} consistently performs well on real-world robotic scenes, including our real-world robot environment, Ego4D~\cite{grauman2022ego4d} and Open X-Embodiment dataset~\cite{o2023open}.
% Additionally, \name{} consistently performs well on real-world datasets, including Ego4D~\cite{grauman2022ego4d} and Open X-Embodiment~\cite{o2023open}, and also facilitates real-world robotic tasks on a real-world robot.
Overall, our contributions are as follows:
\begin{itemize}
    \item We introduce \name{}, an automated, editable video simulation framework that leverages collaborative LLM agents and VLM for robotic manipulation. 
    % allowing users to generate complex simulations using only natural language commands
    \item We present a comprehensive set of simulation operators, featuring well-designed modules such as the ISD module and the 3D-NNFM Loss, to cover diverse scenarios.
    \item We validate the effectiveness of \name{} both quantitatively and qualitatively through extensive experiments conducted on multiple datasets and scenes.
\end{itemize}

\section{Related Work}
\label{sec:related}

\subsection{Robotic Manipulation Simulations}
To advance generalist robot manipulation policies, various physics-based simulation platforms have been developed to provide an efficient and scalable way to expand embodiment data.
Built on physics engines such as Isaac Sim~\cite{isaacsim}, PyBullet~\cite{coumans2019}, and MuJoCo~\cite{todorov2012mujoco}, many robotic manipulation environments and diverse skills have been developed, covering rigid-body manipulation~\cite{bousmalis2023robocat, gervet2023act3d, xian2023chain, xiang2020sapien}, and soft-body manipulation~\cite{weng2022fabricflownet, xian2023fluidlab, wang2023softzoo, seita2023toolflownet} for simulating deformable robots, objects and fluids.
Moreover, some works have also established standardized simulation benchmarking~\cite{li2023behavior, yu2020meta, james2020rlbench, gu2023maniskill2}.
These works significantly reduce reliance on costly real-world data collection.
However, the sim-to-real gap still remains a significant challenge, despite large efforts to address it~\cite{zhuang2023robot, DeXtreme,chang2020sim2real2sim,yuan2024learning, wang2024cyberdemo}.
On the other hand, the Real2Sim2Real approaches seek to address the Sim2Real gap, which remains challenging in transfer for RGB-based manipulation policies from simulation to real world~\cite{20years,robosim2real}.
Several prior works have explored Sim2Real transfer for RGB-based manipulation policies using domain randomization~\cite{exarchos2021policy,huber2024domain} and domain adaptation~\cite{bousmalis2017unsupervised,zheng2023gpdan} techniques. These approaches, however, often require task-specific tuning and environment engineering, which can be both labor-intensive and difficult to achieve accurately in traditional physics simulators.
% More recently, significant progress has been made in the commercialization of robotics. For instance, Hi Robot~\cite{shi2025hi} and Helix~\cite{openai2025Helix} have adopted vision-language-action (VLA) models based on vision-language models (VLMs). However, collecting vision-language data for these models or expanding existing datasets remains a challenge that traditional simulation methods cannot address.
In this paper, we propose \name{}, an automated and editable video simulation framework for robotic manipulation, which enables the construction of photo-realistic, view-consistent simulations with a wide range of simulation operators, allowing users to generate complex simulations using only natural language.

\subsection{Scene Simulation with 3D Gaussian Splating}
Scene reconstruction and simulation have been a longstanding research problem.
Currently, 3D Gaussian Splatting (3DGS)~\cite{kerbl3Dgaussians} has revolutionized this field with explicit representation and high-fidelity real-time rendering, largely extending the capabilities of NeRFs~\cite{mildenhall2021nerf}.
Apart from fast rendering, the explicit representation of 3DGS also
facilitates a range of downstream tasks, including dynamic reconstruction~\cite{wu20234d, yang2023real, cotton2024dynamic}, geometry editing~\cite{GaussianEditor, GaussianEditor_text, GaussCtrl}, physical simulation~\cite{xie2024physgaussian}, and scene understanding~\cite{FeatureSplatting, SemanticGaussians, ye2024gaussian}.
More recently, several studies have explored using 3DGS for robotic manipulation tasks.
For example, GaussianGrasper~\cite{GaussianGrasper} and GraspSplats~\cite{ji2024graspsplats} utilize Feature Splatting~\cite{FeatureSplatting} to support grasp queries via language,
while Robo-GS~\cite{lou2024robo} integrates 3D Gaussian kernels to enhance the digital asset representation of robotic arms.
Although previous works have explored the reconstruction and editing capabilities of 3DGS,  and presented toy demos or simple applications within the robotics domain, 
developing a systematic pipeline to expand the robotic simulation systems and enhance robotic performance, remains an unexplored area.

\section{\name}
\label{sec:method}
% In this section, we introduce the proposed \name{} in detail. We first give an overview of the framework in \cref{subsec: overview}. Then, by considering the misalignment issue, we clarify the formulation of reasoning-decision alignment in \cref{subsec: alignment} and the redesigned CoT in \cref{subsec: CoT}.
\begin{figure*}[ht]
    \centering
    \includegraphics[width=1\linewidth]{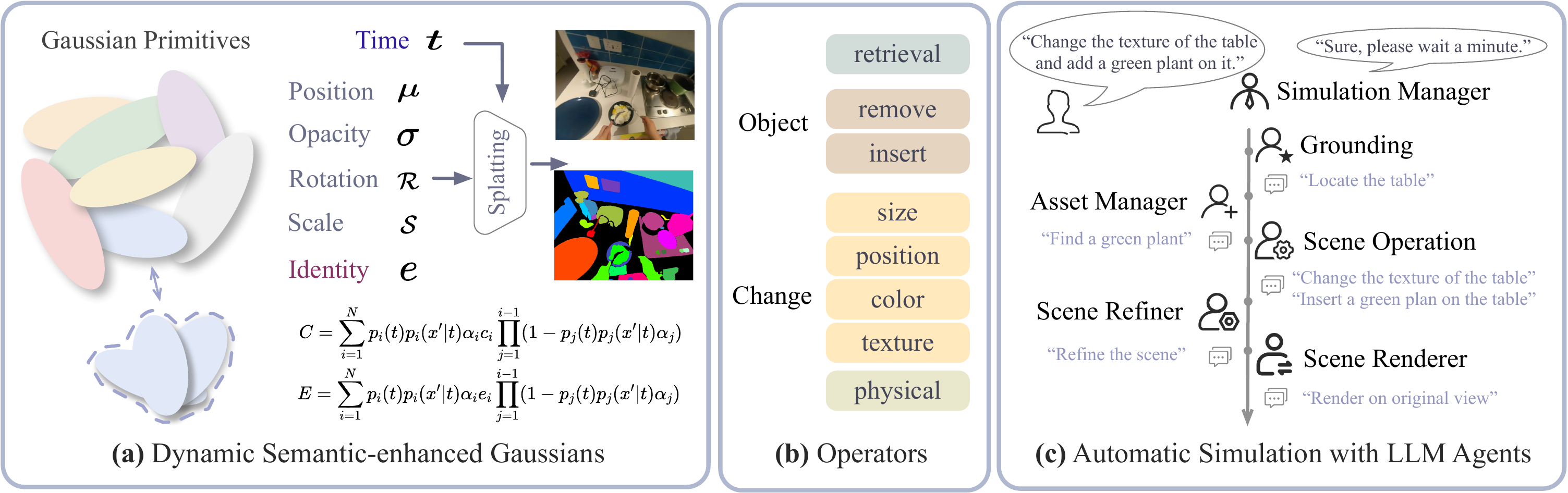}
    \vspace{-15pt}
    \caption{\textbf{(a)} \name{} extends the Gaussian representation to reconstruct dynamic scenes with semantic features from demonstration videos.
    \textbf{(b)} \name{} includes and refines various simulation operators.
    \textbf{(c)} \name{} leverages multiple LLM agents to automate and streamline the simulation production process following user natural language commands. }
    \label{fig:framework}
    \vspace{-10pt}
\end{figure*}
\subsection{Overview}
As illustrated in \cref{fig:framework}, \name{} firstly reconstructs dymanic scenes with semantic features from demonstration videos (\cref{subsec:4dgs}). Then, with various simulation operations (\cref{subsec:simulation}), \name{} leverages LLMs to process user commands (\cref{subsec:llm}).
Furthermore, \name{} utilizes a VLM to enhance robotic performance (\cref{subsec:vlm}).
% Furthermore, \name{} enhances robotic performance by utilizing VLMs to analyze learning issues and generate corresponding simulation demands (\cref{subsec:vlm}).

% \subsection{Constructing Semantic-enhanced Gaussians}
\subsection{Dynamic Semantic-enhanced Gaussians}
\label{subsec:4dgs}

% \qquad \qquad \qquad \qquad \qquad 
% \textcolor{purple}{\textit{\textbf{--- \raisebox{-0.4ex}{\includegraphics[height=1\baselineskip]{imgs/zhenzhushanbei.png}} Oyster to produce pearls}}}

Given demonstration videos, we first reconstruct the scene with Gaussian representation to construct photo-realistic simulations and extend it with temporal and semantic information to enable a wide range of simulation operators.

\mypara{3D Gaussian Splatting.} 
% 太长了，没啥贡献，大家都知道
% 3DGS represents the 3D scene explicitly with multiple Gaussian primitives as $G = \{g_1, g_2, \ldots, g_n\}$. Each Gaussian primitive $g_i$ is parameterized by $\theta_i = (\mu_i, c_i, r_i, s_i, \sigma_i)$ as
% % \begin{equation}
% $
% g_i(x) = e^{-\frac{1}{2}(x-\mu_i)^T \Sigma_i^{-1} (x-\mu_i)},
% $
% % \end{equation}
% where $\mu_i, c_i, r_i, s_i, \sigma_i$ respectively represent the positions, color, rotation, scale, opacity, and $\Sigma_i$ denotes covariance matrix acquired from the rotation and scales as $\Sigma = RSS^TR^T$.
% Then for fast rendering, the 3D Gaussian primitive is projected onto the 2D plane, and the projected 2D covariance matrix in camera coordinates is computed as $\Sigma' = JW \Sigma W^T J^T$, where $J$ is the Jacobian of the affine approximation of the projective transformation and $W$ is the observation matrix.
% The final pixel color $C$ can be rendered by $\alpha$-blending:
% % 3D Gaussian splats that overlap at a given pixel, with the Gaussians sorted in depth order:
% %  a GPU-friendly blending process is used to color the pixels according to a depth-ordered 3D Gaussian:
% \begin{equation}
% \label{eq:3dblending}
% C =\! \sum_{i=1}^{N} p_{i}(x')\alpha_i c_i \prod_{j=1}^{i-1} (1-p_{i}(x')\alpha_j),
% \end{equation}
% where the final opacity $\alpha_i$ is formulated as
% $ \alpha_i = \sigma_i e^{-\frac{1}{2}\left(x' - \mu'_i)^T \Sigma'^{-1}_i (x' - \mu'_i)\right)} $ and $x'$ and $\mu_i'$ are coordinates in the projected space,
% and $p_{i}(x')$ is the probability density of the $i$-th Gaussian at pixel $x'$.
3DGS represents the 3D scene explicitly with multiple Gaussian primitives as $G$. Each Gaussian primitive $g_i$ is parameterized by $\theta_i = (\mu_i, c_i, \Sigma_i, \sigma_i)$ as
% \begin{equation}
$
g_i(x) = e^{-\frac{1}{2}(x-\mu_i)^T \Sigma_i^{-1} (x-\mu_i)},
$
% \end{equation}
where $\mu_i, c_i, \sigma_i$ respectively represent the positions, color, opacity, and $\Sigma_i$ denotes covariance matrix acquired from the rotation and scales as $\Sigma = RSS^TR^T$.
Then the Gaussian primitive is projected onto the 2D plane, with the projected 2D covariance matrix as $\Sigma' = JW \Sigma W^T J^T$, where $J$ is the Jacobian of the projection transformation and $W$ is the observation matrix.
The final pixel color $C$ can be rendered by $\alpha$-blending:
% 3D Gaussian splats that overlap at a given pixel, with the Gaussians sorted in depth order:
%  a GPU-friendly blending process is used to color the pixels according to a depth-ordered 3D Gaussian:
\begin{equation}
\label{eq:3dblending}
C =\! \sum_{i=1}^{N} p_{i}(x')\alpha_i c_i \prod_{j=1}^{i-1} (1-p_{i}(x')\alpha_j),
\end{equation}
where the final opacity $\alpha_i$ is formulated as
$ \alpha_i = \sigma_i e^{-\frac{1}{2}\left(x' - \mu'_i)^T \Sigma'^{-1}_i (x' - \mu'_i)\right)} $ and $x'$ and $\mu_i'$ are coordinates in the projected space,
and $p_{i}(x')$ is the probability density of the $i$-th Gaussian at pixel $x'$.

\mypara{Dynamic Reconstruction.}
% employs a 4D Gaussian parameterized by anisotropic ellipses that can rotate arbitrarily in space and time, as well as view-dependent and time- evolved appearance represented by 4D SH coefficients
The vanilla 3DGS representation, lacking temporal modeling in dynamic scenes, is insufficient for real-world robotic environments. To address this, we enable Gaussian primitives to propagate over time, capturing the spatiotemporal dynamics of the scene.
Specifically, following 4DGS~\cite{yang2023real}, we treat time and spatial dimensions equally to formulate the dynamic Gaussian model by
extending the position to $\mu =(\mu_x, \mu_y, \mu_z, \mu_t)$ and the covariance matrix $\Sigma$ to a 4D ellipsoid,
where $S=\mathrm{diag}(s_x,s_y,s_z,s_t)$  and $R$ is a 4D rotation matrix that can be decomposed into a pair of isotropic rotations.
Subsequently, each frame in a dynamic scene can be represented as a view in a 3D static space, conditioned on a timestamp $t$, and the formula \cref{eq:3dblending} is extended as:
\begin{equation}
\label{eq:4dblending}
    C = \sum^{N}_{i=1}  p_{i}(t) p_{i}(x'|t) \alpha_i c_{i} \prod^{i-1}_{j=1} (1- p_{j}(t) p_{j}(x'|t) \alpha_j).
\end{equation}
% When representing a 4D Gaussian, taking a different perspective on space and time, considering the assumption that $(x,y,z)$ and $t$ are independent of each other, the underlying 4D Gaussian can be achieved by adding an additional one-dimensional Gaussian $pi(t)$ to the original 3D Gaussian, conditional 3D Gaussian can be derived from 4D Gaussian:
% \begin{equation}
% \begin{aligned}
% & \mu_{x y z \mid t}=\mu_{1: 3}+\Sigma_{1: 3,4} \Sigma_{4,4}^{-1}\left(t-\mu_t\right) \\
% & \Sigma_{x y z \mid t}=\Sigma_{1: 3,1: 3}-\Sigma_{1: 3,4} \Sigma_{4,4}^{-1} \Sigma_{4,1: 3}
% \end{aligned}
% \end{equation}
% where $\Sigma$ is the 4D covariance matrix, which is calculated by customized $S_{4d}=diag(s_x,s_y,s_z,s_t)$ and customized $R_{4d} = L\left(q_l\right) R\left(q_r\right)$. $ql = (a, b, c, d)$ and $qr = (p, q, r, s)$ represent left and right isotropic rotation respectively.

% Gaussian Grouping [58] utilizes a 2D identity loss and a 3D regularization loss for Gaussian optimization, leveraging the coherent segmentation across views.

\mypara{Semantic Gaussians.}
% sixteen-dimensional feature vector
To enable scene simulations, we need to decompose the observed scene into distinct components for further manipulation.
% Since many priors~\cite{FeatureSplatting, SemanticGaussians, ye2024gaussian} have explored scene understanding in 3DGS, following Gaussian Grouping~\cite{ye2024gaussian}, 
Building on prior works in 3DGS scene understanding~\cite{FeatureSplatting, SemanticGaussians, ye2024gaussian},
we extend above dynamic Gaussian primitive $g_i$ with a new parameter, identity encoding $e_i$, which is a low-dimensional learnable embedding, allowing the Gaussians to be grouped according to object instance.
The identity encodings are supervised by leveraging the 2D mask predictions by SAM~\cite{kirillov2023segment}, which has demonstrated impressive open-world segmentation capability. 
Similar to rendering the RGB color in \cref{eq:4dblending}, the 2D identity feature $E$ in dynamic scene is rendered as:
\begin{equation}
\label{eq:semantic}
    E = \sum^{N}_{i=1}  p_{i}(t) p_{i}(x'|t) \alpha_i e_{i} \prod^{i-1}_{j=1} (1- p_{j}(t) p_{j}(x'|t) \alpha_j).
\end{equation}
% Then, taking the 2D rendered identity features, 
% Then we apply an extra linear layer $f$ to recover its feature dimension back to $K$, followed by a  $\mathit{softmax}(f(E))$ operation for identity classification, where $K$ is the total number of semantic masks.
Then we apply an extra linear layer $f$ and a  $\mathit{softmax}(f(E))$ function for identity classification.

% It is worth noting that the identity encoding $e$ is independent of timestamp $t$ and remains unchanged throughout the time series.
% Moreover, during densification, newly generated Gaussian primitives inherit the identity encoding of their predecessors. This ensures that Gaussian primitives associated with a specific object do not acquire the identity labels of other objects over time, thereby maintaining spatiotemporal consistency.

\mypara{Overall Optimazition.}
Consequently, we represent the dynamic scene with a set of time-conditioned, grouped Gaussians for manipulation, inheriting SAM’s strong zero-shot scene understanding capability.
For scene reconstruction, we use the MSE rendering loss, $\mathcal{L}_\text{2d}$.
% , as supervision with the ground-truth scene views.
For semantic learning, we adopt a standard cross-entropy loss, $\mathcal{L}_\text{sem}$.
% , for classification across $K$ categories with ground-truth semantic masks.
Following ~\citep{ye2024gaussian}, we also employ a KL divergence loss, $\mathcal{L}_\text{3d}$, to enforce 3D spatial consistency, which constrains the identity encodings of the top $K$-nearest Gaussians to be close in feature space, mitigating the occlusion problem within objects.
The overall optimization objective is:
\begin{equation}
L = \lambda_{2d}L_{2d} + \lambda_{sem}L_{sem}  + \lambda_{3d}L_{3d},
\end{equation}
where $\lambda$ are weight coefficients to balance each loss term.

%%%% 不能写的这么简单的堆积
% \subsection{Automatic Editable Video Simulation}
\subsection{Editable Video Simulation}
\label{subsec:simulation}

% \qquad \qquad \qquad \qquad  
% \textcolor{purple}{\textit{\textbf{--- \raisebox{-0.4ex}{\includegraphics[height=0.9\baselineskip]{imgs/zhenzhu2.png}} Cultured and polished pearls }}}

% Edited scenes serve as valuable training resources for robots, providing a practical alternative when real-world data collection is challenging or time-consuming
% After the training and grouping of dynamic semantic-enhanced Gaussians, we can modify scene elements following user prompts to achieve desired simulations.
% Thanks to the decoupled scene representation, instead of fine-tuning all 3D Gaussians, we only need to adjust a small subset of existing or newly added 3D Gaussians relevant to the editing targets, significantly improving simulation efficiency.

After the training and grouping of dynamic semantic-enhanced Gaussians, we refine and polish various simulation operators to cover diverse scenarios. Detailed operator frameworks are provided in the Appendix.
% Thanks to the decoupled scene representation, instead of fine-tuning all 3D Gaussians, we only need to adjust a small subset 3D Gaussians relevant to the editing targets, significantly improving simulation efficiency.

\mypara{Incremental Object Retrieval.} 
To perform simulations based on user commands, the first step is to retrieve the target object.
Since SAM does not directly support language prompts, we adopt the G-DINO~\cite{liu2023grounding} to identify the desired 2D object and obtain its corresponding mask ID. This mask ID is then matched with our rendered segmentation masks.
However, due to the infinite granularity of objects based on user needs, such as retrieving small buttons on a stove, the 2D masks used for training may not cover these fine-grained object parts, leading to retrieval failures.
While recent methods~\cite{li2024langsurf, peng2024gags} attempt to use SAM's multi-level masks to represent small, medium, and large object hierarchies, this still cannot represent all object granularities and increases the training cost of semantic-enhanced Gaussians.
To address this issue, we propose Incremental Semantic Distillation (ISD) to incrementally distill object semantics into the scene.
Specifically, upon retrieving the desired object Gaussians, we render the 2D object mask and use G-DINO to verify whether it corresponds to the desired object, such as the small button, or mistakenly retrieves the entire object, such as the stove. 
If the target object is not identified, we further use bounding boxes as prompts to SAM for a finer-grained segmentation.
Then, we fine-tune only the identity encoding $e$ of the previously retrieved 3D object Gaussians with the new fine-grained labels, thereby incrementally distilling the new object semantic. Since only the identity encoding of relevant targets needs to be updated, the entire process remains highly efficient.

\mypara{Object Removal.} 
3D object removal can be achieved by simply deleting the 3D object Gaussians. However, in real-life videos captured from a few perspectives, removing an object may leave behind a blurry hole in the background due to insufficient observations. To address this, we first detect the ``blurry hole" using G-DINO~\cite{liu2023grounding} and apply LAMA~\cite{lama} inpainting on each view. Then, we generate new Gaussians near the deletion area and fine-tune only these newly introduced Gaussians using the inpainted views, ensuring a seamless reconstruction of the missing background.

\mypara{Object Insertion.}
3D object insertion can be achieved by inserting the corresponding 3D Gaussians. However, several challenges need to be addressed.
The first challenge is obtaining the Gaussian representation of the desired object. Fortunately, large-scale datasets of Gaussians, i.e., ShapeSplat\cite{ma2024shapesplat} and uCO3D\cite{liu2025uncommon}, already contain a vast range of common and uncommon objects,
which can serve as our initial database, enabling direct retrieval and use.
% Additionally, prior works have explored efficient 3D generation through just a feed-forward process, such as GRM\cite{xu2024grm} and LGM\cite{tang2024lgm}. These methods enable the generation of 3D Gaussian representations from textual descriptions or single-view images, providing a flexible way to acquire the required objects.
Additionally, prior works have explored efficient 3D generation from textual descriptions or single-view images through just a feed-forward process, such as GRM\cite{xu2024grm} and LGM\cite{tang2024lgm}, providing a flexible way to acquire the required objects.
The second challenge arises when inserting objects from external sources. Even after adjusting the position and size appropriately, the inserted object may still exhibit noticeable color contrast with the original scene, making it appear unrealistic.
To address this issue, we adopt libcom~\cite{niu2021making}, a comprehensive image composition library that encompasses various algorithms such as image blending and standard/painterly image harmonization, to refine the rendered image to ensure color consistency between the surrounding scene and the newly inserted object.
Subsequently, we fine-tune the spherical harmonic (SH) of the inserted object Gaussians with the refined rendered image, which typically takes only a few minutes.

\mypara{Object Modification.}
We support various modifications.
% We also support various fine-grained 3D object modifications.

\noindent \textit{\textbf{-- Size and Position.}}
For 3D object size modification, we adjust the properties of the target Gaussians by scaling them accordingly. 
In practice, $\mu_i$ is directly scaled, and $r_i$ and $s_i$ need to be scaled in the logarithmic space using additive adjustments.
3D object position modification is a combination of object removal and object insertion. Specifically, we first remove the target object from its original position and then insert it into the desired location.

\noindent \textit{\textbf{-- Color.}}
For 3D object color modification, we adjust the spherical harmonic (SH) of the object Gaussians to preserve the learned 3D scene geometry.
However, simply changing the color can lead to severe distortions, as it does not account for lightness variations. 
To this end, we carefully adopt the CIELAB color space, which enables color modification while preserving the original lighting effects.

% By rendering images from the 3D scene and editing them using Instruct Pix2Pix (IPix2Pix) [2], IN2N iteratively updates the 3D scene until convergence. As there is no guarantee of consistent editing of multi-view images, this method suffers from instability, slow processing speeds, and notable artefacts,
\noindent \textit{\textbf{-- Style and Texture.}}
For 3D object style/texture modification, several works~\cite{GaussianEditor, GaussianEditor_text, GaussCtrl} explore to leverage diffusion models (e.g., IP2P~\cite{brooks2022instructpix2pix}) to iteratively edit the rendered images while updating the 3D reconstruction.
However, image diffusion models do not inherently enforce multi-view consistency, which is critical for preventing artifacts.
% In practice, these methods often fail to restrict modifications to the target object, unintentionally altering undesired areas. 
Moreover, the iterative optimization process is very slow, requiring many full 3D optimizations, making training time-consuming and difficult to control.
To effectively incorporate the desired texture into rendered images while ensuring multi-view consistency, we adopt the nearest neighbor feature matching (NNFM) loss from ARF~\cite{zhang2022arf}.
Specifically, given a rendered image and a reference image, we extract their features from VGG16 ($F_r$ and $F_t$) and then minimize the cosine distance between the feature of each pixel in the rendered image with its nearest neighbor in the reference image: $
L_{\text{NNFM}} = \frac{1}{N}\sum_{i} \min_{j} \left( F_r(i) , F_t(j) \right)
$, where $N$ is the number of pixels in the rendered image.
 % primarily concentrate on stylizing the whole scene with NeRF representation.
Although the vanilla NNFM loss effectively transfers complex high-frequency visual details into the 3D scene, it's limited to stylizing the entire scene.
We extend it to a 3D regularized NNFM loss by 1) optimizing only the SH parameters of the target 3D object Gaussians to preserve the background's spatial details, and 2) regularizing the optimization with the original reconstruction loss to further prevent artifacts caused by SH refinement at object boundaries.  
In practice, we render masks of the target 3D object ($M_{3d}$) to apply the NNFM loss while enforcing the reconstruction loss on regions outside the mask:
\begin{equation}
    % L_{\text{3DNNFM}} = \frac{1}{N}\sum_{i} \min_{j} \left( M_{3d} F_r(i) , F_s(j) \right)
    % + L_{gs}.
   L_{\text{3D-NNFM}}  =   L_{\text{NNFM}}^{M_{3d}} + L_{\text{gs}}^{\overline{{M}_{3d}}}.
\end{equation}

\mypara{Physics Simulation.}
Leveraging our well-reconstructed Gaussian representation, we also enable physics simulation by integrating physical properties into the Gaussian primitive.
The physical parameters include material density ($\rho$), Young's modulus ($E$), and Poisson's ratio ($P$).
Following PhysGaussian~\cite{xie2024physgaussian}, we can manually set physical parameters to target objects and predict their motion using a physical simulator, Material Point Method (MPM)~\cite{hu2018moving}.
Furthermore, to reduce the reliance on manual parameter assignment, we incorporate GPT-4V~\cite{openai2023gpt4v} alongside a material library~\cite{xu2024gaussianproperty} to automatically assign the corresponding physical properties to enhance the physics simulation.

% We can modify the position of objects in the scene, and copy objects in the scene infinitely, such as ``move the cup to the left" and ``spread the cup across the entire desktop".
% The LLM is responsible for understanding the received commands while the role functions process the received data. 
% Each agent is equipped with unique LLM prompts and role functions tailored to their specific duties within the system. 
%  This workflow endows agents with both language interpretation capabilities and precise execution capabilities.
% Tailored LLM prompts ensure that each agent functions optimally and produces the desired outcomes effectively.
% This cohesive operation allows the system to efficiently create and interact complex 3D environments. 
\subsection{Automatic Simulation with LLM-Agents}
\label{subsec:llm}

% \qquad \qquad \qquad \qquad \qquad 
% \textcolor{purple}{\textit{\textbf{--- \raisebox{-0.7ex}{\includegraphics[height=1.1\baselineskip]{imgs/xl2 (1).png}}  String pearls to necklace}}}

Previously, we meticulously designed and refined various simulation operators. To connect these components, we leverage LLM agents as the ``string," automating and streamlining the simulation production process. 
However, directly applying a single LLM agent struggles with multi-step reasoning and cross-referencing multiple operators. 
To address this, we deploy multiple collaborative LLM agents, each equipped with unique prompts and tailored toolsets.
% to specific tasks.
Specifically, each agent first interprets and converts user simulation commands into structured configurations with its specialized LLM prompts, then invokes the corresponding toolsets to generate the desired simulations.
Following, we outline the involved agents:

% The simulation manager agent enhances the system’s robustness in interpreting various inputs and streamlines operations for clarity and fine granularity.
\mypara{Simulation Manager Agent.}
This agent serves as the team leader, decomposing user commands into simplified, concrete natural language instructions and dispatching tasks to other agents.
To enable command decomposition, we design a series of prompts for its LLM. The core idea of the prompts is to describe the simulation operator set, specify the overall goal, and define the output form with examples.

% To achieve this, we design non-trivial LLM prompts (detailed in the Appendix) that interpret user commands $T_{user}$ into a set of structured instructions $T_{a}$, each specifying an editing action assigned to a corresponding agent to accomplish the intended simulation.

\mypara{Grounding Agent.}
This agent processes prompts in the form of \textit{Locate $\langle$object$\rangle$} and employs the object retrieval operator. The returned outcomes include the mask ID and the position of the target object Gaussians. 
For straightforward queries such as ``the red cup," the agent directly retrieves the corresponding object. For more complex queries like ``the cup closest to the pressure cooker," the LLM performs a multi-step process: it first identifies both the pressure cooker and the cups and then determines the spatial relationship between them to retrieve the desired object.

\mypara{Scene Operation Agent.}
The scene operation Agent processes manipulation prompts such as \textit{Change color of $\langle$object$\rangle$} and applies the appropriate simulation operators proposed in \cref{subsec:simulation} to achieve the desired simulation.

\mypara{3D Asset Management Agent.}
This agent organizes and retrieves 3D assets based on user specifications. First, it utilizes an LLM to interpret user commands and retrieves specific 3D objects from our Gaussian database by matching object attributes such as color and type.
If the matching is unavailable, the agent employs an object Gaussian generation model to synthesize the desired object, which is then incorporated into the database to facilitate scalability and versatility. The database and efficient generation process are discussed in the Object Insertion section of \cref{subsec:simulation}.

\mypara{Scene Refiner Agent.}
This agent enhances the overall simulation quality. Since each agent operates independently, cumulative edits may degrade scene reconstruction quality. Thus, a final refinement step is applied to all modified Gaussians to ensure realism and coherence across the simulation.

\mypara{Scene Renderer Agent.}
This agent generates appropriate extrinsic camera parameters for rendering. First, it leverages an LLM to interpret user viewpoint adjustment instructions into relative camera parameters, based on the original viewpoint’s position and orientation.
It then returns the simulated images with the desired perspectives.
% to meet specific user or application needs.

\mypara{Oveall Workflow.}
All tailored agents collaborate to execute the simulation based on user commands following a sequential pipeline:
The simulation manager agent leads the process by dispatching instructions to appropriate agents. 
First, the grounding agent identifies the relevant objects and locations. 
Next, the scene operation agent performs the specified modifications, optionally assisted by the 3D asset management agent if new assets are required. 
Then, the scene refiner agent enhances the overall simulation quality to ensure consistency and realism. 
Finally, the scene renderer agent generates the desired video output to return.
Detailed agents' prompts and toolset are shown in the Appendix.
% Additionally, the Simulation Manager Agent records all simulation configurations, enabling multi-round editing and further refinements as needed.

% GPT-4~\cite{achiam2023gpt4}

% Deepseek-VL~\cite{lu2024deepseekvl}
% GPT-4V~\cite{openai2023gpt4v}
% \section{Collaboration with Robot Learning}
% \subsection{Efficient Robot Learning with \name{}}
% identifies reason about
\subsection{Efficient Robot Learning}
\label{subsec:vlm}

% \qquad \qquad \qquad \qquad
% \textcolor{purple}{\textit{\textbf{--- \raisebox{-0.4ex}{\includegraphics[height=1\baselineskip]{imgs/kou2.png}} Clasp to link pearl necklace}}}

Despite being trained on increasingly large datasets, robotic models often struggle in specific environments or datasets, requiring human experts to proactively identify failures and retrieve or collect additional cases to enhance robot training.
While our proposed automatic simulation framework helps mitigate data collection challenges, current approaches still rely heavily on human expertise, limiting the model's learning and evolution. 
While some prior works~\cite{du2023vision, sagar2024robofail}, such as RoboFail~\cite{sagar2024robofail}, have explored failure detection, they primarily treat failure reasoning as a binary classification problem, lacking in-depth failure analysis.
To overcome these limitations, we take a further step by leveraging the recent, knowledgeable Vision-Language Model (VLM) to replace human experts in reasoning about robotic manipulation failures.
% using natural language. 
Specifically, we utilize a VLM to analyze keyframes of failure cases and provide detailed explanations across potential failure causes, such as object position, color, and background texture.
% In practice, we employ two VLMs, i.e., GPT-4V~\cite{openai2023gpt4v} and Deepseek-VL~\cite{lu2024deepseekvl} to cross-validate and ensure the accuracy.
After identifying the specific failure cause, we further prompt the VLM to generate corresponding simulation solutions in natural language. These textual instructions are then fed into our proposed automatic simulation framework, which generates targeted simulations to enhance model training.
Automatically identifying issues and generating simulation commands closes the simulation loop, like a “clasp” to link the pearl necklace, to form our complete method, \name{}, ultimately driving more efficient and robust robot learning.
% By automatically identifying issues and generating simulation commands, we not only close the simulation loop but also complete our RoboPearls method—like a clasp securing the final link of a pearl necklace—ultimately driving more efficient and robust robot learning.

% The model assesses the potential failure reasons (foreground, background, or weather) three times and outputs the confidence for each category. The prompt used for this process is shown in the part (a) of Figure~\ref{fig:prompt_1}. A threshold of 0.8 is set; if the confidence surpasses this threshold, the VLM is further instructed to provide the specific cause under the identified failure category (foreground, background, or weather). 

\section{Experiment}
\label{sec:exp}
% In this section, we first introduce the experimental setup in \cref{subsec:setup}. Then, the main results are present in \cref{subsec:mainres}. The analysis of each component is present in \cref{subsec:ablation}. 

We present the experimental setup in \cref{subsec:setup}, main results in \cref{subsec:mainres}, and ablation analysis in \cref{subsec:ablation}. 

\subsection{Experimental Setting}
\label{subsec:setup}
We conduct experiments on multiple datasets and scenes, i.e., RLBench~\cite{james2020rlbench}, COLOSSEUM~\cite{pumacay2024colosseum}, Ego4D~\cite{grauman2022ego4d}, Open X-Embodiment~\cite{o2023open}, and a real-world robot. Detailed implementations are in the Appendix.

\subsection{Main Results}
\label{subsec:mainres}
\begin{table*}[ht]
\centering \footnotesize
\caption{\textbf{Results on Colosseum}. \name{} demonstrates significant performance improvements across all perturbations.
}
\vspace{-3pt}
\addtolength{\tabcolsep}{-2.6pt}
\begin{tabularx}{\linewidth}{lcccccccccc}

% SAM2Act~\cite{fang2025sam2act}           &       \textbf{-4.3}$\pm$3.6           &                     \textbf{-1.1}$\pm$2.5       &-\textbf{0.7}$\pm$7.2   &  \textbf{-3.3}$\pm$2.4    &  \textbf{24.72}$\pm$6.1    &    -15.9$\pm$5.0     &  \textbf{0.9}$\pm$6.8      &        \\

\toprule
\textbf{Method}     & \textbf{Avg. Success $\uparrow$} & \textbf{MO-Color} & \textbf{RO-Color} & \textbf{MO-Texture} & \textbf{RO-Texture} & \textbf{MO-Size} & \textbf{RO-Size}  \\
\midrule
RVT~\cite{goyal2023rvt}  & 51.7 $\pm$ 3.8 &  52.4 $\pm$ 3.3 & 48.8 $\pm$ 4.1 & 39.0 $\pm$ 4.8 & 59.3 $\pm$ 3.8 & 75.6 $\pm$ 2.5 & 55.9 $\pm$ 3.6   \\
RVT-2~\cite{goyal2024rvt2}    &  64.6 $\pm$ 4.7 &  64.1 $\pm$ 5.3 & 70.6 $\pm$ 4.7 & 58.1 $\pm$ 6.6 & 68.3 $\pm$ 4.4 & 81.7 $\pm$ 3.7 & 67.9 $\pm$ 4.6   \\

% SAM2Act~\cite{fang2025sam2act} &         \\
\midrule

\rowcolor{mygray} \bf \name{}-RVT (Ours)  & \bf69.2 $\pm$ 3.4 &\bf  67.6 $\pm$ 2.7 &\bf 71.7 $\pm$ 3.4 & \bf66.0 $\pm$ 4.9 & \bf70.9 $\pm$ 3.9 & \bf83.3 $\pm$ 3.2 & \bf68.0 $\pm$ 3.8 \\
\rowcolor{mygray} \bf \name{}-RVT2 (Ours) & \bf75.4 $\pm$ 3.5 &\bf  76.5 $\pm$ 3.3 &\bf 80.9 $\pm$ 4.9 & \bf73.8 $\pm$ 3.5 & \bf77.8 $\pm$ 3.6 & \bf86.8 $\pm$ 3.4 & \bf74.9 $\pm$ 4.4\\
% \rowcolor{mygray} \bf \name{}-SAM2Act (Ours) \\

\toprule
\toprule
\textbf{Method}     & \textbf{Light Color} & \textbf{Table Color} & \textbf{Table Texture} & \textbf{Distractor} & \textbf{Background Texture} & \textbf{Camera Pose} & \textbf{All Perturbations}  \\
\midrule
RVT~\cite{goyal2023rvt}  & 49.8 $\pm$ 3.8 & 49.5 $\pm$ 4.0 &  48.8 $\pm$ 4.5 & 59.4 $\pm$ 3.9 & 58.2 $\pm$ 3.9 & 57.5 $\pm$ 3.5 & 17.8 $\pm$ 3.8\\
RVT-2 \cite{goyal2024rvt2}     &  62.1 $\pm$ 5.7 & 59.1 $\pm$ 4.3 &  61.7 $\pm$ 4.3 & 68.1 $\pm$ 3.7 & 73.1 $\pm$ 3.7 & 68.5 $\pm$ 5.1 & 36.1 $\pm$ 4.5   \\
% SAM2Act~\cite{fang2025sam2act}   &      \\

\midrule

\rowcolor{mygray} \bf \name{}-RVT (Ours)  & \bf70.9 $\pm$  3.0 &\bf 72.0 $\pm$ 3.3 & \bf 69.7 $\pm$ 3.3 &\bf 66.3 $\pm$ 3.1 & \bf74.1 $\pm$ 3.1 &\bf 68.2 $\pm$ 3.3 & \bf 50.8 $\pm$ 3.6 \\
\rowcolor{mygray} \bf \name{}-RVT2 (Ours) & \bf75.0 $\pm$  3.2 &\bf 77.8 $\pm$ 3.0 & \bf 74.8 $\pm$ 2.9 &\bf 71.0 $\pm$ 3.1 & \bf80.0 $\pm$ 3.0 &\bf 77.1 $\pm$ 2.6 & \bf 54.7 $\pm$ 4.6\\
% \rowcolor{mygray} \bf \name{}-SAM2Act (Ours) \\

\bottomrule
\end{tabularx}
\label{tab:colosseum}
% \vspace{-5pt}
\end{table*}

\begin{table*}[ht]
\centering \footnotesize
\caption{\textbf{Results on RLBench.}  \name{} achieves remarkable performance gains over state-of-the-art models.}
\vspace{-3pt}
\addtolength{\tabcolsep}{-1.6pt}
\begin{tabularx}{\linewidth}{lccccccc}
\toprule

% \textbf{Method}     & \textbf{Avg. Success $\uparrow$} & \textbf{Avg. Rank $\downarrow$} & \textbf{Close Jar} & \textbf{Drag Stick} & \textbf{Insert Peg} & \textbf{Meat off Grill} & \textbf{Open Drawer} & \textbf{Place Cups} & \textbf{Place Wine} & \textbf{Push Buttons} \\

% \textbf{Method}     & \textbf{Put in Cupboard} & \textbf{Put in Drawer} & \textbf{Put in Safe} & \textbf{Screw Bulb} & \textbf{Slide Block} & \textbf{Sort Shape} & \textbf{Stack Blocks} & \textbf{Stack Cups} & \textbf{Sweep to Dustpan} & \textbf{Turn Tap} \\

% RVT~\cite{goyal2023rvt}                & 62.9 $\pm$ 3.7                   & 3.6                               & 52.0 $\pm$ 2.5     & 99.2 $\pm$ 1.6      & 11.2 $\pm$ 3.0      & 88.0 $\pm$ 2.5          & 71.2 $\pm$ 6.9       & 4.0 $\pm$ 2.5       & 91.0 $\pm$ 5.2      & \textbf{100.0} $\pm$ 0.0       \\

\textbf{Method}     & \textbf{Avg. Success $\uparrow$} &  \textbf{Stack Cups} & \textbf{Push Buttons} & \textbf{Insert Peg} & \textbf{Put in Cupboard} & \textbf{Basketball in Hoop} & \textbf{Close Box} \\

\midrule

RVT~\cite{goyal2023rvt}                &  62.4 $\pm$ 3.0 & 17.6 $\pm$ 4.1 & 97.6 $\pm$ 1.9 & 16.4 $\pm$ 2.1 & 50.4 $\pm$ 3.2 & 98.4 $\pm$ 1.9 & 94.4 $\pm$ 3.2 \\
RVT-2~\cite{goyal2024rvt2}               &  70.1 $\pm$ 4.6 & 51.0 $\pm$ 8.2 & 98.0 $\pm$ 2.1 & 26.5 $\pm$ 6.0 & 52.5 $\pm$ 7.5 & 98.0 $\pm$ 2.1 & 94.5 $\pm$ 2.1      \\
SAM2Act~\cite{fang2025sam2act}    &   83.8 $\pm$ 3.5 & 63.2 $\pm$ 4.6 & 100.0 $\pm$ 0.0 & 88.0 $\pm$ 5.4 & 60.8 $\pm$ 4.6 & 98.4 $\pm$ 3.2 & 92.8 $\pm$ 3.0     \\
\midrule
\rowcolor{mygray} \bf \name{}-RVT (Ours) & \bf  68.0 $\pm$ 2.3 &\bf 28.4 $\pm$ 3.8 & \bf \textcolor{mypurple}{100 $\pm$ 0.0}& \bf24.0 $\pm$ 2.8 & \bf 60.8 $\pm$ 3.9 & \bf100.0 $\pm$ 0.0 & \bf 94.8 $\pm$ 3.6\\
\rowcolor{mygray} \bf \name{}-RVT2 (Ours) & \bf  78.0 $\pm$ 4.7 &\bf 67.4 $\pm$ 3.6 & \bf \textcolor{mypurple}{98.3 $\pm$ 3.1}& \bf33.7 $\pm$ 7.6 & \bf 75.5 $\pm$ 7.8 & \bf98.0 $\pm$ 2.1 & \bf 95.0 $\pm$ 3.1\\
\rowcolor{mygray} \bf \name{}-SAM2Act (Ours) & \bf  88.5 $\pm$ 2.4 &\bf 68.0 $\pm$ 4.0 & \bf \textcolor{mypurple}{100.0 $\pm$ 0.0}& \bf93.6 $\pm$ 3.7 & \bf 72.1 $\pm$ 3.9 & \bf100.0 $\pm$ 0.0 & \bf 97.3 $\pm$ 3.0\\
\bottomrule
\end{tabularx}% 

\label{tab:rlbench}
\vspace{-3pt}

\end{table*}

\begin{figure*}[h!]
    \centering
    \includegraphics[width=1\linewidth]{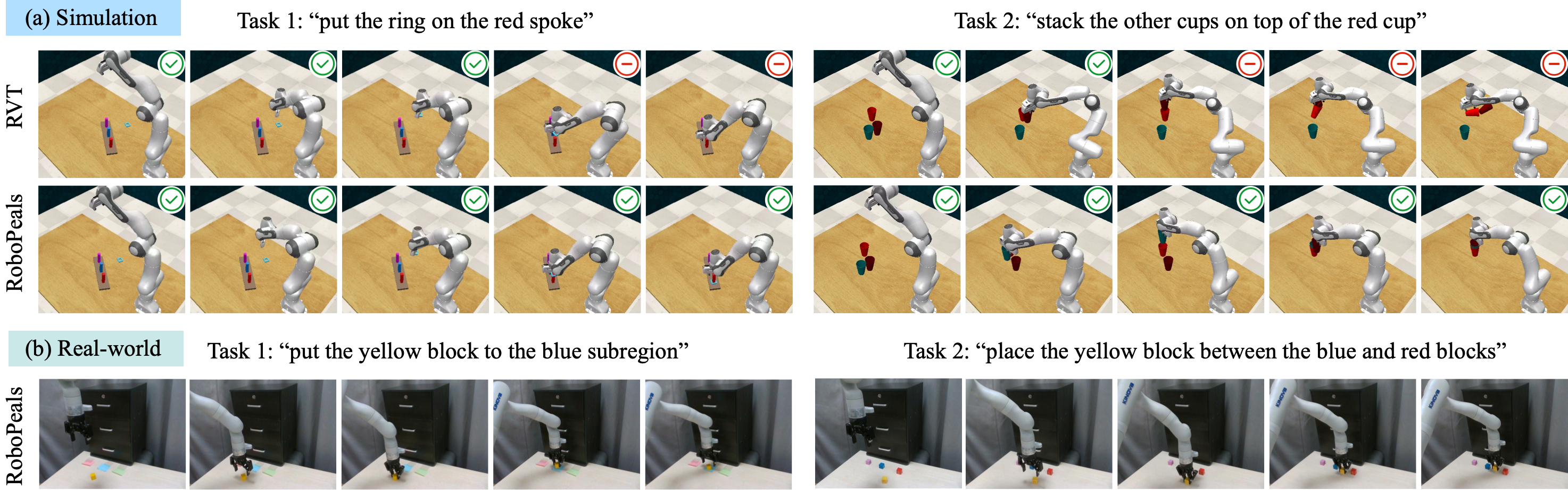}
    % \vspace{-15pt}
    \caption{\textbf{The demonstrations for manipulation tasks in simulation (a) and the real world (b)} (zoom-in for the best of views).
    % The red mark denotes that the pose deviates severely from the demonstration, while the green mark indicates that the pose aligns with the ground-truth trajectory. 
    } 
    \label{fig:resrl}
   % \vspace{-10pt}
   \vspace{-3pt}
\end{figure*}

\mypara{Results on simulation datasets.}
% The Colosseum benchmark and tested them under 13 different perturbation categories over three runs.
In \cref{tab:colosseum}, we present task completion success rates for 13 perturbations on COLOSSEUM, providing a systematic evaluation of robustness under various conditions. Our approach, \name{}, demonstrates significant performance improvements across all perturbations, with average success gains of 17.5\% and 10.8\% over RVT and RVT2, showing its robustness against environmental variations (e.g., lighting changes) and object-level perturbations (e.g., color changes). 
Additionally, in \cref{tab:rlbench}, we report results on several challenging tasks from RLBench to assess general manipulation performance. 
Overall, \name{} achieves average success rates of 68.0\%, 78.0\%, and 88.5\%, boosting the baseline models by 5.6\%, 7.9\%, and 4.7\%, respectively. 
These improvements across benchmarks highlight the efficiency of our editable simulation framework, which leverages various operators to handle diverse scenarios effectively.
% highlighting the effectiveness of our photo-realistic, view-consistent simulations.

As illustrated in \cref{fig:resrl} (a), we present two qualitative examples of the generated action sequence. In the right case, the agent is instructed to “stack the other cups on top of the red cup”. The results indicate that the previous agent struggles to complete the task, whereas our \name{} precisely identifies each cup and successfully stacks them onto the red one. This improvement is attributed to our method’s efficient simulation, which enables manipulation models to have a more accurate understanding of diverse scenes.
% “stack the other cups on top of the red cup”
% “put the ring on the red spoke”

\begin{table}[t]
  \centering \footnotesize
   \caption{\textbf{Results on the real-world robot.} We evaluate each model $20$ times with seen/unseen objects.}
  % \small
  \vspace{-4pt}
   \addtolength{\tabcolsep}{-3.6pt}
    \begin{tabularx}{\linewidth}{l|cccccc}
\toprule
 \multirow{2}{*}{\textbf{Method}} &\multicolumn{2}{c}{\textbf{Pick up}} & \multicolumn{2}{c}{\textbf{Put on}} & \multicolumn{2}{c}{\textbf{Place in}}    \\
 \cmidrule(r){2-7}& Seen & Unseen & Seen & Unseen  & Seen & Unseen\\
\midrule
% RDT~\cite{liu2024rdt} & 4 / 20 & 0 / 20 & 0 / 20 & 0 / 20 & 2 / 20 & 1 / 20  \\
% \midrule
RDT~\cite{liu2024rdt} & 10 / 20 & 4 / 20 & 7 / 20 & 0 / 20 & 8 / 20 & 1 / 20  \\
\rowcolor{mygray} \bf \name{} (Ours)& \bf 15 / 20 &\bf 14 / 20 &\bf 10 / 20 &\bf 9 / 20 &\bf 12 / 20 & \bf 12 / 20  \\

\bottomrule
% \multicolumn{7}{l}{\footnotesize{We evaluate each model $20$ times with seen/unseen objects.} }
\end{tabularx}
 
\vspace{-5pt}
\label{tab: real}
\end{table}

\begin{figure*}[ht]
    \centering
    \includegraphics[width=1\linewidth]{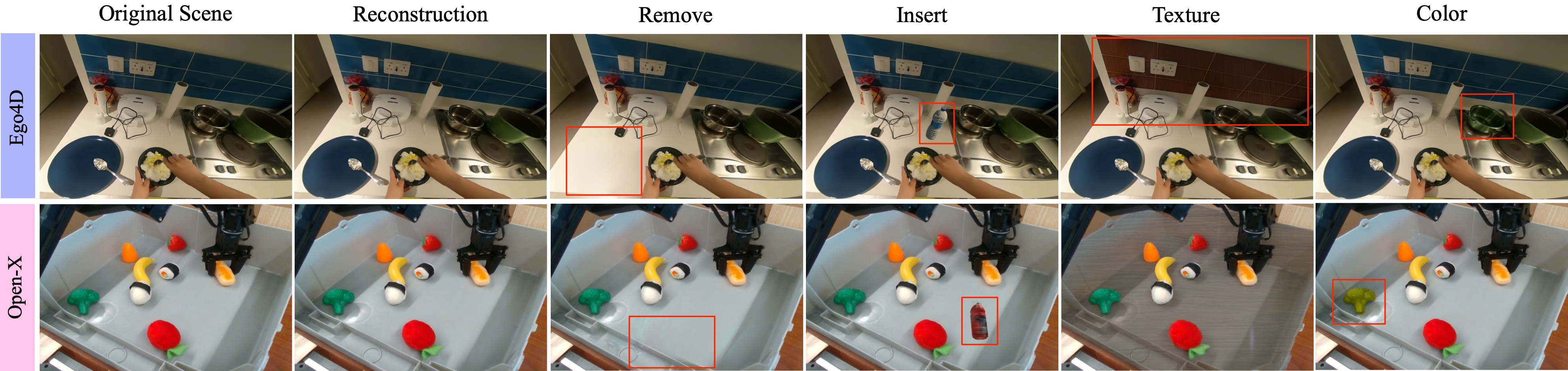}
    \caption{\textbf{The photo-realistic simulations on in-the-wild datasets.} Our \name{} supports various simulations.}
    \label{fig:res}
    \vspace{-4pt}
\end{figure*}

% 去掉RLbench的simulation 到 supp
\mypara{Results on real-world robot.}
In \cref{tab: real} and \cref{fig:resrl} (b), we evaluate the performance of \name{} on a real-world robotic system, which is built on the Kinova Gen3 robotic arm.
% Our experimental setup consists of a 7-DoF Kinova Gen3 robotic arm paired with a Robotiq 2F-85 gripper. 
% We utilized a Kinova Gen3 ultra-lightweight robotic arm to collect RGB-D frame-action data for six tasks.
% We set up both a third-person camera and a wrist camera.
As illustrated, \name{} successfully performs real-world tasks with remarkable generalization ability, whereas the baseline struggles, verifying our effectiveness in real-world environments. Please refer to the Appendix and videos for more details on the setup and performance.
% RDT~\cite{liu2024rdt}
% The baseline agent struggles to complete the task, while our \name{} returns to the red square and successfully slides the square to the yellow target, owing to that our method with efficient simulation can correctly understand the diverse scene, which is consistent with the results in the simulation benchmark.

\mypara{Results on real-world datasets.}
In \cref{fig:res}, \name{} consistently achieves photo-realistic, view-consistent simulations with various operators on real-world datasets, including the Ego4D and the Open X-Embodiment dataset.

\subsection{Ablation Study}
\label{subsec:ablation}
We conduct quantitative and qualitative ablations to comprehensively evaluate our designed modules' effects. 

\begin{table}[ht]
  \centering \footnotesize
   \caption{\textbf{Quantitative ablations on the proposed modules.}}
  % \small
  \vspace{-3pt}
   \addtolength{\tabcolsep}{-2.5pt}
    \begin{tabularx}{\linewidth}{l|ccc}
\toprule
\textbf{Method} & \textbf{Stack Cups} & \textbf{Put in Cupboard} & \textbf{Insert Peg}  \\
\midrule
RVT~\cite{goyal2023rvt} & 14.5 $\pm$ 3.9 & 40.4 $\pm$ 4.0 & 11.0 $\pm$ 3.0               \\
+ IP2P~\cite{brooks2022instructpix2pix} &  18.4 $\pm$ 3.9  & 44.8 $\pm$ 4.9 & 10.7 $\pm$ 3.8\\
% RVT-2~\cite{goyal2024rvt2}                &        \\
% SAM2Act~\cite{fang2025sam2act}    &       \\
\midrule
\name{} (w/o VLM) & 24.7 $\pm$ 3.4 & 45.0 $\pm$ 4.9 & 16.5 $\pm$ 4.1\\
\rowcolor{mygray} \bf  \name{} (Ours) & \bf 37.7 $\pm$ 4.6 &  \bf 55.5 $\pm$ 4.5 &  \bf 17.1 $\pm$ 5.3 \\

% \begin{table}[t]
%   \centering \footnotesize
%    \caption{\textbf{Quantitative ablations on the proposed modules.}}
%   % \small
%    \addtolength{\tabcolsep}{3pt}
%     \begin{tabularx}{\linewidth}{l|c}
% \toprule
% Method & Avg. Success $\uparrow$   \\
% \midrule
% RVT~\cite{goyal2023rvt}                &  23.5 $\pm$ 3.5 \\
% + InstructPix2Pix~\cite{brooks2022instructpix2pix} & 26.5 $\pm$ 4.3\\
% % RVT-2~\cite{goyal2024rvt2}                &        \\
% % SAM2Act~\cite{fang2025sam2act}    &       \\
% \midrule
% \name{}-RVT (w/o VLM) & 30.5 $\pm$ 3.2\\
% \name{}-RVT (Ours) \\

% HSSD  & color-space & VLM & Avg. Success $\uparrow$   \\
% \midrule

%  % \no  & \yes  & \yes  & \yes   & \yes   & \yes  &\cellcolor{ImportantColor}0.4846    \\
%  %   \yes  & \no  & \yes  & \yes   & \yes   & \yes  &\cellcolor{ImportantColor}0.4846    \\
%  %    \yes  & \yes  &\no  & \yes   & \yes   & \yes  &\cellcolor{ImportantColor}0.4846    \\
%  %     \yes  & \yes  & \yes   &\no  & \yes   & \yes  &\cellcolor{ImportantColor}0.4846    \\
%  %     \yes  & \yes  & \yes   & \yes   & \no  &  \yes  &\cellcolor{ImportantColor}0.4846    \\
%  %       \yes  & \yes  & \yes   & \yes   & \yes &  \no   &\cellcolor{ImportantColor}0.4846    \\
%  % \midrule
%  %    \yes  & \yes  & \yes   & \yes   & \yes &  \yes   &\cellcolor{ImportantColor}0.4846    \\

% \colorbox{mycyan}{\cmark} &    \xmarkg &  \xmarkg &0.4846    \\
%  \cmarkg  &   \colorbox{mycyan}{\cmark} &  \xmarkg & 0.4846    \\
%   \midrule
%  \cmarkg  &  \cmarkg  & \colorbox{mycyan}{\cmark}   & 0.4846    \\

\bottomrule
\end{tabularx}
 
\vspace{-3pt}
\label{tab: ablation}
\end{table}

\begin{figure}[ht]
    \centering
    \includegraphics[width=1\linewidth]{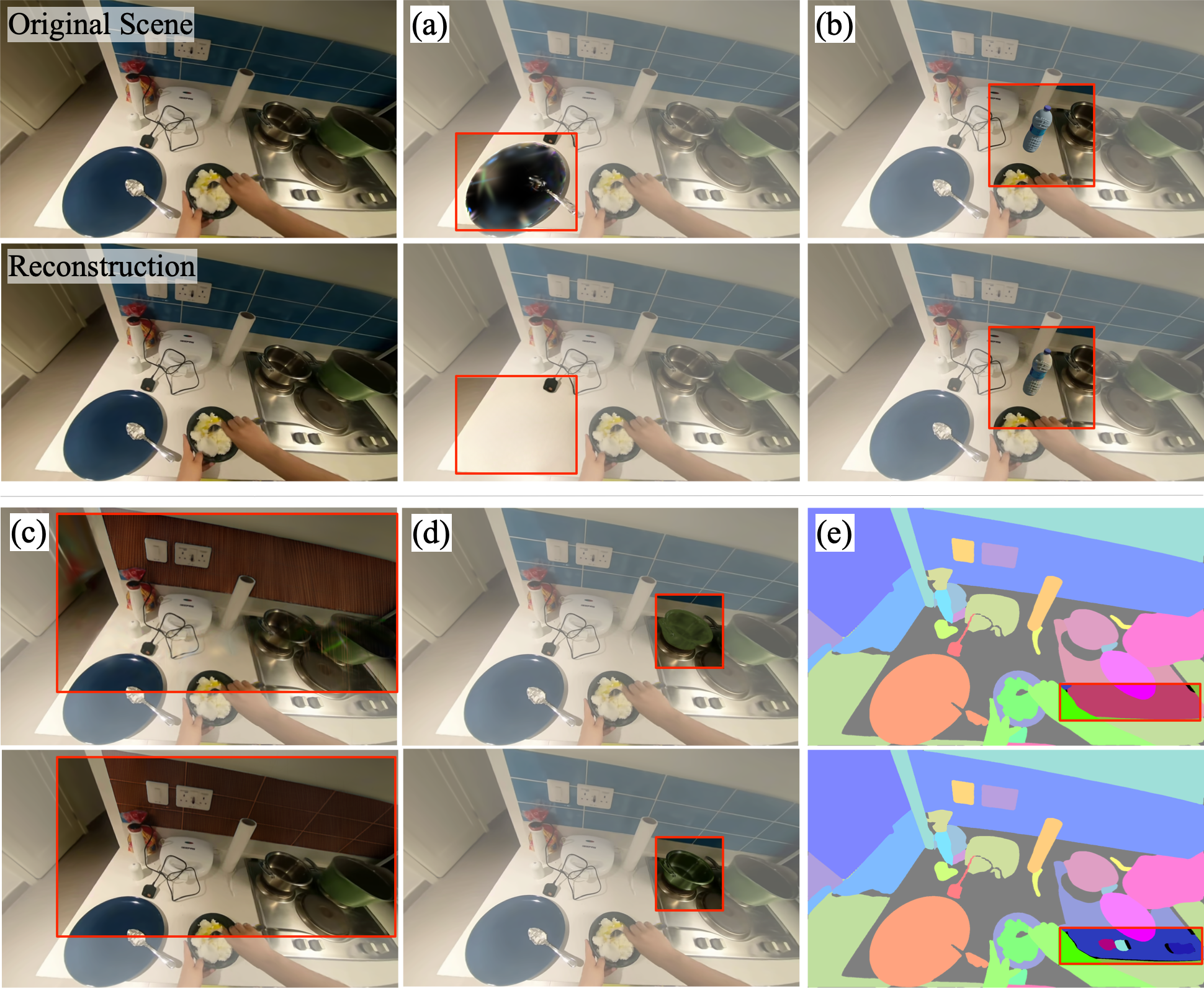}
    %\vspace{-15pt}
    \caption{\textbf{Qualitative ablations on the proposed modules}
    % (zoom-in for the best of views).
    }
    \label{fig:ablation}
 \vspace{-5pt}
\end{figure}

\mypara{Quantitative ablations.}
In \cref{tab: ablation}, we first validate the view-consistent 3D simulation capability of \name{} by comparing it with IP2P~\cite{brooks2022instructpix2pix}, which can be regarded as simulators on 2D image space.
While IP2P provides limited improvement in some cases, our significantly superior results demonstrate the effectiveness of our simulation with spatial-temporal consistency. 
Additionally, we also verify the impact of incorporating VLM, which efficiently analyzes learning issues and generates corresponding simulation demands, leading to enhanced robotic performance.

\mypara{Qualitative ablations.}
In \cref{fig:ablation}, we systematically ablate key designs on the simulation operators, including (a) direct deletion vs. inpainting and fine-tuning, (b) direct insertion vs. refinement with libcom and fine-tuning, (c) NNFM loss vs. 3D regularized NNFM loss, (d) RGB space vs. CIELAB color space, and (e) direct semantic distillation vs. incremental semantic distillation.
The visual results demonstrate that all our designs effectively contribute to constructing photo-photorealistic simulations.

% \subsection{Qualitative Results}
% \label{subsec:vis}
% \input{latex/fig/res}

\section{Conclusion}
In this paper, we introduce \name{}, an automated editable video simulation framework for robotic manipulation. 
Leveraging Gaussian representations, \name{} generates highly adaptable and photorealistic simulations from demonstration videos. 
Moreover, \name{} supports a wide range of simulation operators to cover various scenarios, driven by well-designed modules such as Incremental Semantic Distillation and 3D regularized NNFM Loss. 
To further streamline the process, \name{} integrates LLMs and VLM, allowing users to generate complex simulations using only natural language commands while enabling advanced closed-loop simulation capabilities. 
These features facilitate robust simulations for diverse robotic tasks. Extensive experiments across multiple datasets demonstrate the framework's simulation effectiveness, yielding significant improvements in robotic performance.
Overall, \name{} represents a significant step toward providing a scalable, user-friendly solution for robotic simulation.

%, offering both flexibility and efficiency.

% In this paper, we introduce \name{}, an automated editable video simulation framework to enhance robotic manipulation.
% By leveraging 3DGS, \name{} allows for the creation of photo-realistic and highly adaptable simulations from demonstration videos. 
% Through the integration of large language models (LLMs) and vision-language models (VLMs), we enable users to generate complex simulations using only natural language demands, significantly reducing the need for human intervention while enabling advanced close-loop simulation capabilities. 
% The wide range of simulation operators, supported by well-designed modules such as HSSD and 3D-NNFM Loss, equips \name{} to handle a diverse set of scenarios and provides robust simulations for various robotic tasks. 
% Our extensive experiments on multiple datasets demonstrate the framework’s effectiveness, yielding substantial improvements in robotic performance.
% Overall, \name{} represents a significant step toward offering a scalable, user-friendly solution for robotic manipulation tasks. 
% 感觉句子直接不够连续，读起来费劲

% % Future work will focus on further improving the framework’s flexibility, extending its capabilities, and exploring its application to more complex real-world robotic systems.

\section*{Acknowledgments}
This work is supported by National Key Research and Development Program of China (2024YFE0203100), National Natural Science Foundation of China (NSFC) under Grants No.62476293, Shenzhen Science and Technology Program No.GJHZ20220913142600001, Nansha Key R\&D Program under Grant No.2022ZD014, and General Embodied AI Center of Sun Yat-sen University.

% \clearpage
% \newpage
{
    \small
    \bibliographystyle{ieeenat_fullname}
    \bibliography{main}
}


\clearpage

% phantomsection helps with hyperlinks and references, when appendix is placed
% after the bibliography.
\newpage
\appendix
\phantomsection

% \twocolumn[
% \begin{@twocolumnfalse}
% \section*{\centering \name{}: Unified Multimodal Sensor Generation for Autonomous Driving\\ Supplementary Materials}
% \addcontentsline{toc}{section}{Supplementary Materials}
% \end{@twocolumnfalse}
% ]
% \clearpage
% \setcounter{page}{1}
\maketitlesupplementary

% Begin local group for supplementary material
\begingroup

% Adjust the depth of the table of contents
\setcounter{tocdepth}{2}

% Start a new table of contents for the supplementary material
% \renewcommand{\contentsname}{Contents}
\startcontents[supplements]

% Redefine the section and subsection numbering within the appendix
\renewcommand{\thesection}{A.\arabic{section}}
\renewcommand{\thesubsection}{\thesection.\arabic{subsection}}

% Reset the section counter
\setcounter{section}{0}

% Print table of contents for the supplementary material
% \printcontents[supplements]{}{1}{}
% Contexts
% \hypersetup{linkbordercolor=black,linkcolor=black}
% \hypersetup{linkbordercolor=red,linkcolor=red}

\setlength{\cftbeforesecskip}{0.8em} % 上下距离
\cftsetindents{section}{0em}{2.5em}
\cftsetindents{subsection}{1em}{3.3em}

\etoctoccontentsline{part}{Appendix}
\localtableofcontents

% \clearpage
% \setcounter{page}{1}
% \maketitlesupplementary

% \section{Limitations and Future work}
\section{Social Impact and Limitations.}
\subsection{Future Work}
As this paper is the first to construct photo-realistic simulations, there remain many benefits yet to be explored.
\begin{itemize}
\item First, our approach inherently supports novel view synthesis. While RVT~\cite{goyal2023rvt} leverages point clouds to generate virtual images,
our method enables the rendering of realistic images from new viewpoints, potentially unlocking significant advancements for models like RVT.
More recently, Act3D~\cite{gervet2023act3d} and SAM2Act~\cite{fang2025sam2act} have leveraged foundation models, such as CLIP and SAM, to extract image embeddings for robotic manipulation. However, these single-image foundation models lack a fundamental understanding of the 3D scene and are sensitive to view variations, leading to the generation of noisy and view-inconsistent masks or features, which in turn cause manipulation failures.
In contrast, our \name{} distills open-world semantics into 3D space and can render view-consistent semantic features, as shown in \cref{fig:feature}, offering the potential to further improve the performance of recent sota models.
\item Second, the interaction perspective remains largely unexplored. Existing methods, such as GraspSplats~\cite{ji2024graspsplats}, use 3DGS to reconstruct static scenes, whereas our approach handles dynamic environments with semantic features, allowing for more effective generation of grasping proposals using explicit Gaussian primitives.
\item Third, recent vision-language-action (VLA) models based on VLMs, such as Hi Robot~\cite{shi2025hi} and Helix~\cite{openai2025Helix}, have garnered significant attention for their ability to process complex instructions and generate dexterous manipulations. 
However, collecting and scaling vision-language data remains a significant challenge.
Our \name{} framework allows users to generate complex simulations using natural language instructions, which enables the effective expansion of vision-language-aligned datasets and potentially enhances model performance.
\item Fourth, our work advances the goal of realistic closed-loop simulation. Currently, we utilize VLM to enhance training performance in a closed-loop manner. However, our high-quality reconstruction and rendering capabilities enable robust closed-loop policy evaluation, and pave the way for developing a GS-based closed-loop reinforcement learning (RL) training paradigm to further improve robotic learning.
\end{itemize}

\subsection{Limitaions}
Beyond these promising directions, there are also limitations that we aim to address in future work. 
\begin{itemize}
\item The primary challenge lies in generalization, as our approach relies on scene-specific training for each environment. Fortunately, recent advancements in Generalizable Gaussian Splats~\cite{charatan2024pixelsplat, chen2024mvsplat} provide a promising avenue to mitigate this issue. 
\item Moreover, as shown in \cref{tab: time}, the training and processing time for high-resolution real-world scenes is still somewhat lengthy. Fortunately, recent methods~\cite{zhao2024scaling, lu2024turbo} have significantly optimized and accelerated Gaussian fitting, offering the potential for further improvements.
\item Additionally, while incorporating VLM reduces reliance on human experts to some extent, handling highly complex scenes remains challenging. A practical approach is to first utilize VLM to enhance model performance and only involve human intervention when necessary. This strategy is still an improvement over previous methods that relied entirely on human analysis.
\end{itemize}

\begin{table}[ht]
  \centering \footnotesize
   \caption{\textbf{Processing time of each module.} The time is evaluated on Ego4d real-world scene with 120 frames on the resolution of $1918\times1237$ on 1 GPU.}
  % \small
  \vspace{-2pt}
   \addtolength{\tabcolsep}{-1pt}
    \begin{tabularx}{\linewidth}{l|cccccc}
\toprule
Module & Recon & Insert  & Remove & Color & Texture & Physics \\
\midrule
Time (min) & $ \sim $70 & $ \sim $6 & $ \sim $6 & $ \sim $1 & $ \sim $5 & $ \sim $7               \\

\bottomrule
\end{tabularx}
\vspace{-4pt}
\label{tab: time}
\end{table}

\subsection{Scope}
RoboPearls supports various scene edits (\cref{fig:framework} (b) and \cref{subsec:simulation}), except for action-relevant edits, more exactly, generating new manipulation trajectories, which is out of scope. We leave action editing for future work.

Altogether, we hope that our work will inspire further research and contribute to the advancement of robotic simulation and learning.

\section{Additional Related Work}
\subsection{LLM Agent}

Significant advances in Large Language Models (LLMs), e.g., GPT-4~\cite{achiam2023gpt4} and DeepSeek-R1~\cite{guo2025deepseekr1}, have demonstrated their remarkable capabilities across various domains. 
By integrating LLMs as agents, many works~\cite{mao2023language, yang2023llmgrounderopenvocabulary3dvisual, qin2024diffusiongpt} have enhanced problem-solving abilities in interactive and autonomous applications. For example, 
% LLM-Grounder~\cite{yang2023llmgrounderopenvocabulary3dvisual} employs an LLM as a proxy to improve grounding capabilities, while
AutoGen~\cite{wu2023autogen} leverages well-organized LLM agents to form operating procedures and code programming.
ChatSim~\cite{wei2024editable} adopts an LLM-agent collaboration workflow for editing 3D driving scenes, and RoboGen~\cite{wang2023robogen} uses generative models and LLMs to generate robotic tasks.
In this work, we utilize LLM agents to decompose user simulation demands into concrete commands for specific editing functions, thereby automating and streamlining the simulation process.

\section{Additional Details}
\label{sec:details}

\subsection{More Details on Dynamic Gaussians}
To integrate semantic information into the dynamic Gaussians, it is worth noting that the identity encoding $e$ is independent of timestamp $t$ and remains unchanged throughout the time series.
Moreover, during densification, newly generated Gaussian primitives inherit the identity encoding of their predecessors. This ensures that Gaussian primitives associated with a specific object do not acquire the identity labels of other objects over time, thereby maintaining spatiotemporal consistency.

\subsection{More Details on Editable Video Simulation}
\begin{figure}[ht]
    \centering
    \includegraphics[width=1\linewidth]{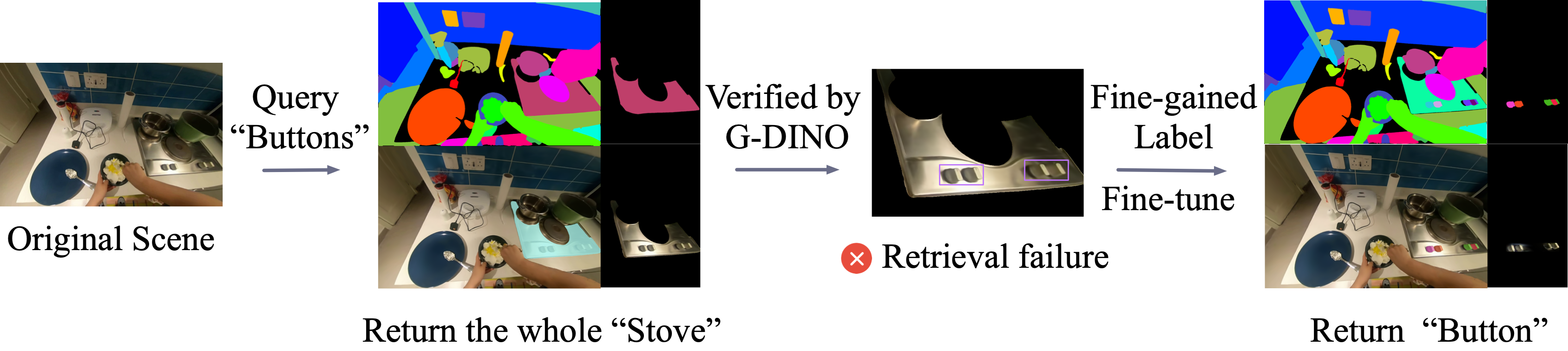}
    % \vspace{-15pt}
    \caption{\textbf{The incremental semantic distillation pipeline} (zoom-in for the best of views).
}
    \label{fig:semantic}
    % \vspace{-10pt}
\end{figure}
\mypara{Incremental Semantic Distillation.}
The pipeline is in \cref{fig:semantic}. After retrieving the desired object Gaussians, we render the 2D object mask and use G-DINO to verify whether it corresponds to the desired object. 
If the target object is not identified, we further use bounding boxes as prompts to SAM for a finer-grained segmentation and fine-tune the identity encoding of the retrieved object Gaussians.

\begin{figure}[ht]
    \centering
    \includegraphics[width=1\linewidth]{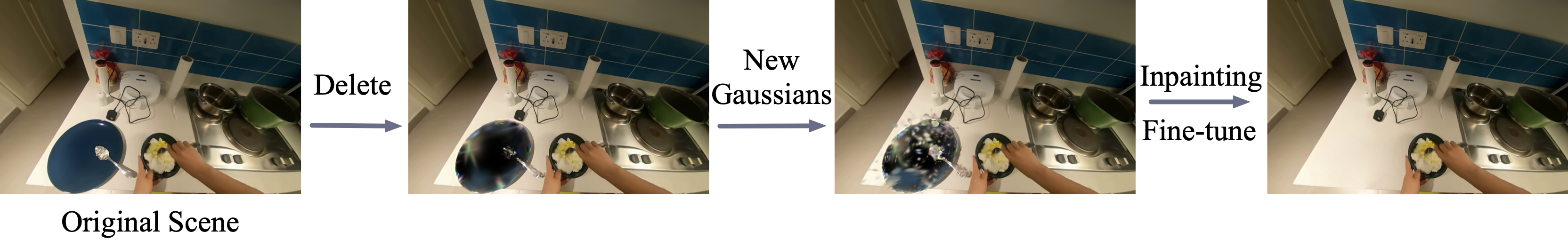}
    % \vspace{-15pt}
    \caption{\textbf{The object removal pipeline.} 
}
    \label{fig:remove}
    % \vspace{-10pt}
\end{figure}

\mypara{Object Removal.} The pipeline is shown in \cref{fig:remove}.
After deleting the 3D object Gaussians, we use G-DINO to detect the ``blurry hole" and apply LAMA inpainting on each view. Then, we generate new Gaussians near the deletion area and fine-tune only these newly introduced Gaussians with the inpainted views.

\begin{figure}[ht]
    \centering
    \includegraphics[width=1\linewidth]{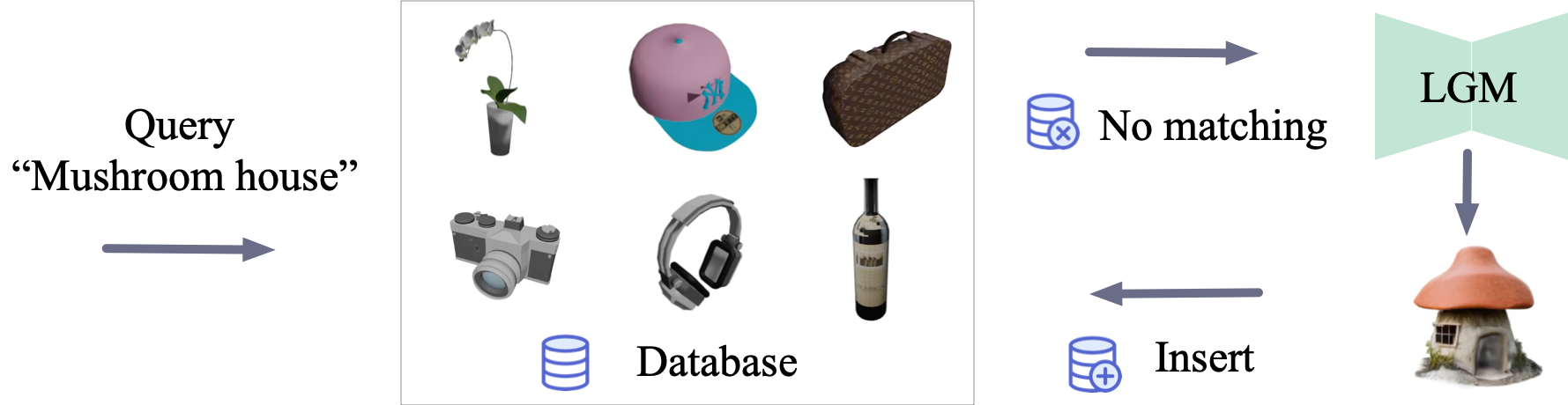}
    % \vspace{-15pt}
    \caption{\textbf{The 3D asset management pipeline.} 
}
    \label{fig:insert}
    % \vspace{-10pt}
\end{figure}
\mypara{3D Asset Management.} The pipeline is shown in \cref{fig:insert}.
The 3D Asset Management agent first retrieves desired 3D objects from the Gaussian database by matching object attributes such as color and type.
If the matching is unavailable, the agent employs a Gaussian object generation model, e.g., LGM, to synthesize the desired object, and then incorporate it into the database. 

\begin{figure}[ht]
    \centering
    \includegraphics[width=1\linewidth]{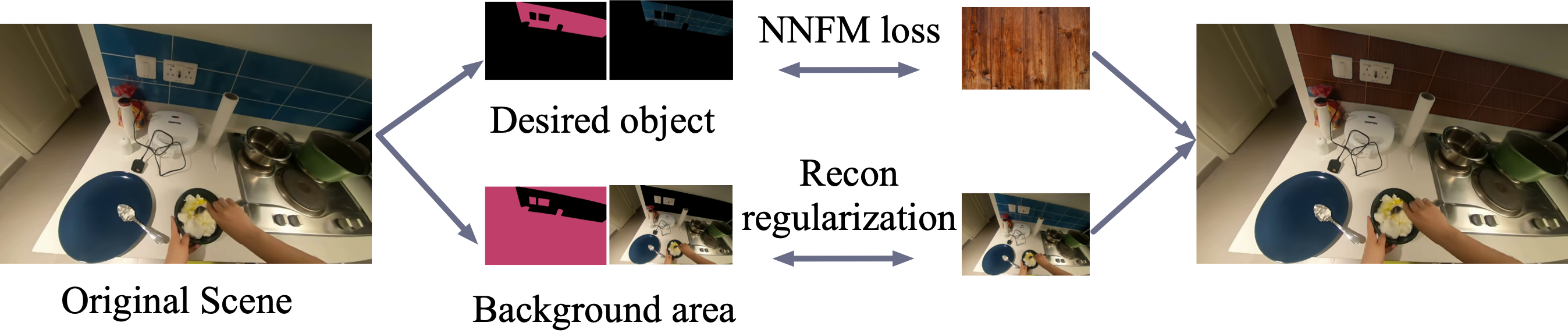}
    % \vspace{-15pt}
    \caption{\textbf{The texture modification pipeline} (zoom-in for the best of views). 
}
    \label{fig:texture}
    % \vspace{-10pt}
\end{figure}
\mypara{Texture Modification.} The pipeline is shown in \cref{fig:texture}.
We render masks of the target 3D object to optimize only the SH parameters of the object Gaussians with the NNFM loss 
while preventing artifacts on regions outside the mask with the original reconstruction loss.

\subsection{More Details on the LLM Agents}
\begin{figure*}[ht]
    \centering
    \includegraphics[width=1\linewidth]{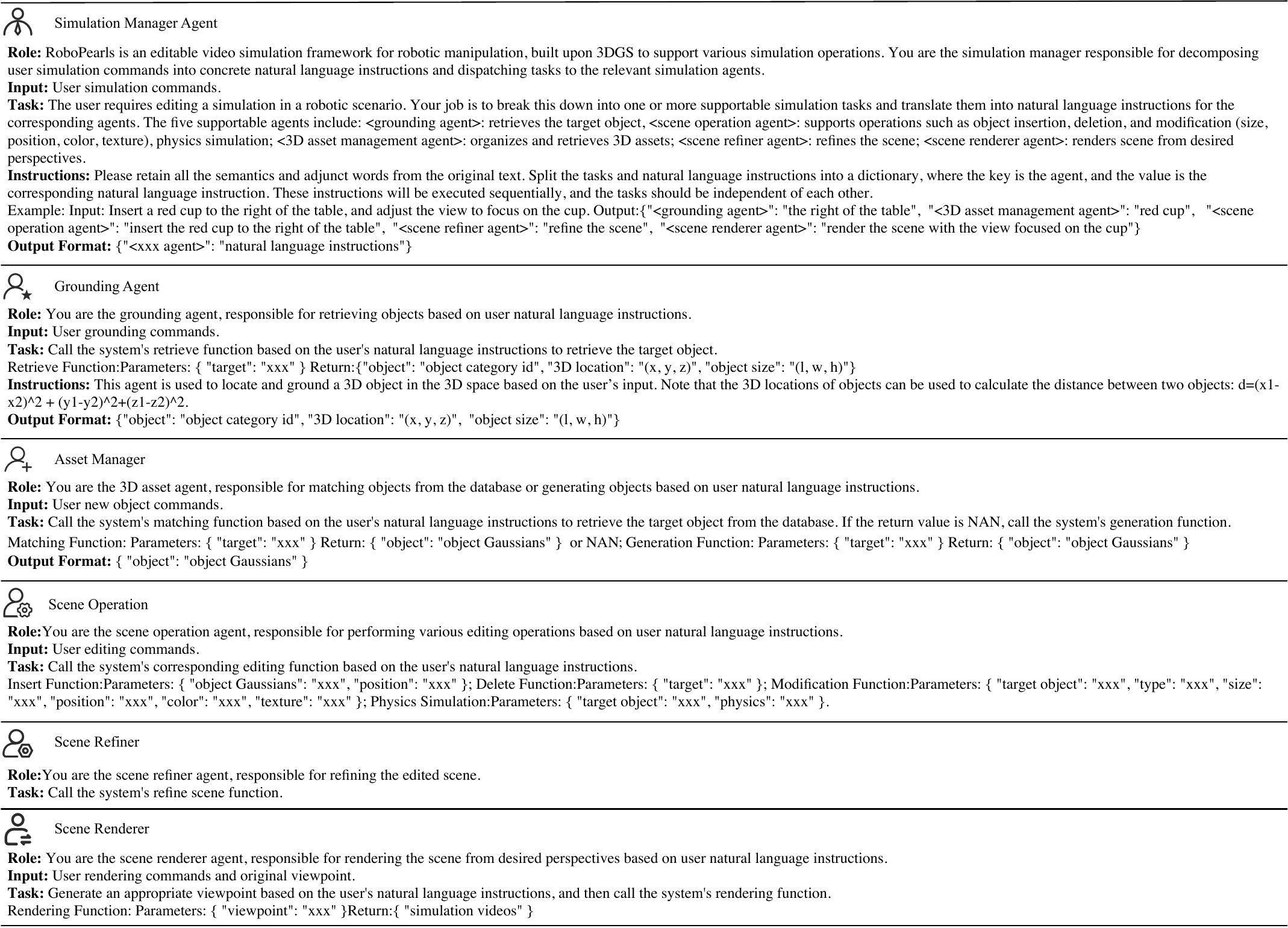}
    % \vspace{-15pt}
    \caption{\textbf{The prompts and examples of the LLM agents} (zoom-in for the best of views).  
}
    \label{fig:llm}
    % \vspace{-10pt}
\end{figure*}
In \cref{fig:llm}, we provide detailed prompts and examples of the LLM agents.
Each agent is equipped with unique LLM prompts and tailored system functions to their specific duties. 
All tailored agents collaborate to execute the simulation based on user commands following a sequential pipeline. Additionally, the Simulation Manager Agent records all simulation configurations, enabling multi-round editing and further refinements as needed.

\subsection{More Details on the VLM}
\begin{figure*}[ht]
    \centering
    \includegraphics[width=1\linewidth]{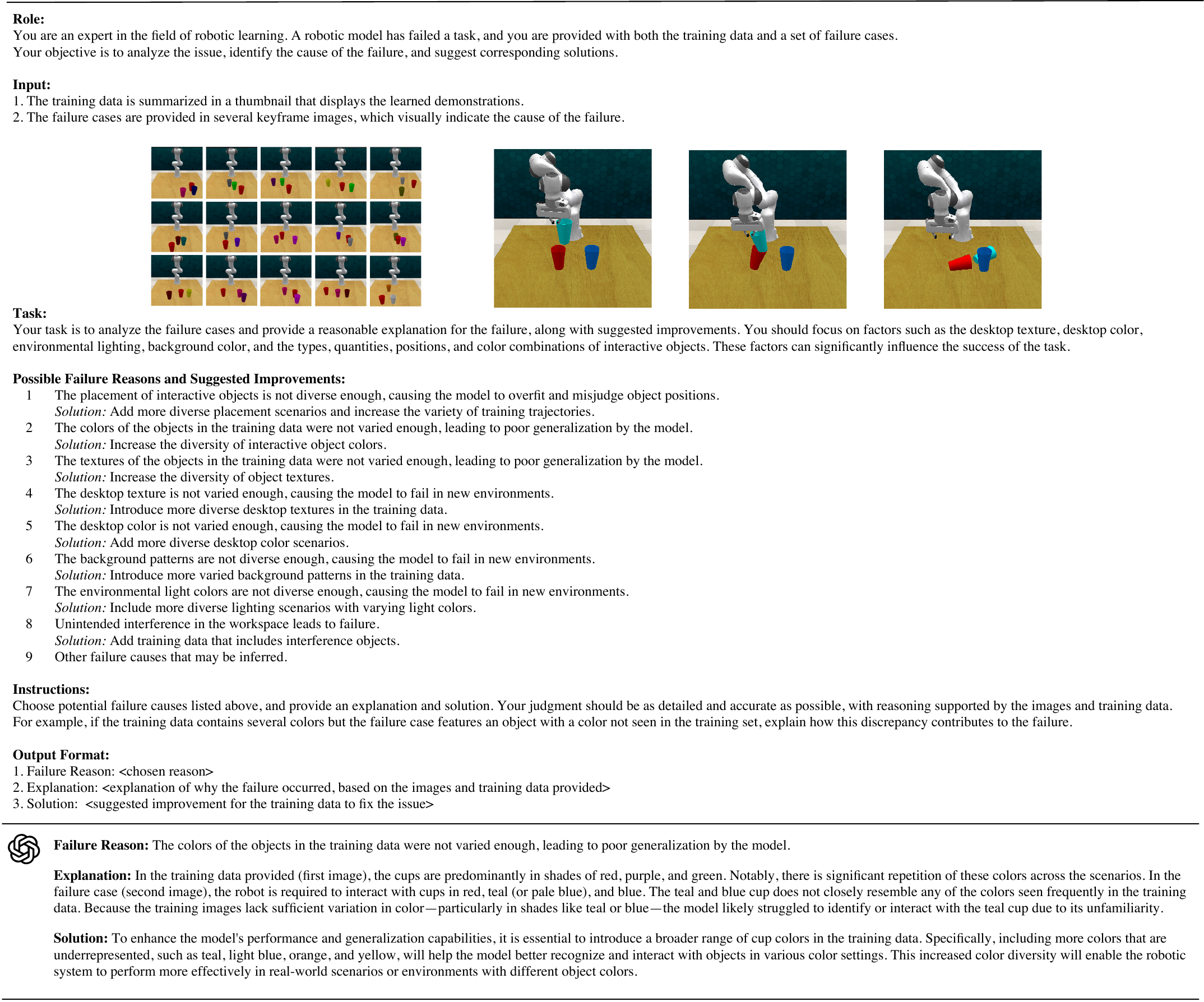}
    % \vspace{-15pt}
    \caption{\textbf{The prompts and examples of the VLM} (zoom-in for the best of views).  
}
    \label{fig:vlm}
    % \vspace{-10pt}
\end{figure*}
In \cref{fig:vlm}, we provide detailed prompts and examples of the VLM.
The VLM is prompted to analyze keyframes of failure cases and provide detailed explanations across potential failure causes, such as object position, color, and background texture. Then the VLM further generates corresponding simulation solutions in natural language, which are then fed into our simulation framework to generate targeted simulations for enhancing model training.

\subsection{Experimental Setting}
\label{subsec:setup_supp}
% ManiGaussian ECCV 2024 只做了RLbench
\mypara{Datasets and Simulation Setup.}
Our experiments are conducted in the popular multi-task manipulation benchmark RLBench~\cite{james2020rlbench} and the generalization evaluation benchmark COLOSSEUM~\cite{pumacay2024colosseum}. 
RLBench is built on the CoppelaSim~\cite{rohmer2013v} simulator, where a Franka Panda Robot is controlled to manipulate the scene. 
% The benchmark contains $18$ tasks, including non-prehensile tasks like \textit{push buttons}, and common pick-and-place tasks like \textit{place wine}. 
The benchmark contains 100 tasks, including picking and placing, high-accuracy peg insertions.
Each task is specified by a language description and consists of $2$ to $60$ variations, which concern scene variability across object poses, appearance, and semantics. 
% such as handling objects in different colors or locations. 
% The diversity of these tasks requires the model to not just mimic the provided expert demonstrations to specialize in one specific skill but rather learn generalizable knowledge about different manipulation skills. 
There are four RGB-D cameras positioned at the front, wrist, left shoulder, and right shoulder. % The wrist camera moves during manipulation.  
The evaluation metric is the task completion success rate, which is the proportion of execution trajectories that achieve the goal conditions specified in the language instructions~\cite{gervet2023act3d, shridhar2023perceiver}.
% The evaluation metric is the task success rate, which measures the percentage of completed episodes. An episode is considered successful if the agent completes the goal specified in natural language within a maximum of \num{25} steps.
COLOSSEUM is a benchmark for evaluating the generalization of robotic manipulation. 
It introduces 13 perturbations across 20 different tasks from the RLBench framework such as \textit{close box}, and \textit{basketball in hoop}. 
These perturbations include changes in color, texture, size of objects and backgrounds, lightnings, distractors, and camera poses. 
For simulation, due to limited computing resources, we utilize a curated subset of 6 challenging language-conditioned manipulation tasks, and follow the original COLOSSEUM setup and generate 13 environmental perturbations for each task. 
% The objects that can be changed include Manipulation object (MO), Receiver Object (RO) and the table. 
% Therefore, it is well-suited for evaluating the generalization ability of manipulation approaches with pertaining.
% For simulation, we follow the original COLOSSEUM setup and generate 14 environmental perturbations for each task. 
% For each environmental perturbation, we generate 25 demonstrations.
We also report the task completion success rate on COLOSSEUM. 
% Instead of reporting the average success rate for each individual tasks, we report the average success rate for each pertubation of the environment, as it will highlight how each method is robust to different pertubations.

\mypara{Baselines.}
We compare \name{} with various models that have been specifically designed for 3D object manipulation including RVT~\cite{goyal2023rvt}, RVT2~\cite{goyal2024rvt2}, and SAM2Act~\cite{fang2025sam2act}, which are the previous SOTA methods on RLBench and COLOSSEUM. 
Since SAM2Act~\cite{fang2025sam2act} does not provide the code and model for COLOSSEUM, we only conducted comparisons with it on RLBench.

\mypara{Implementation Details.}
Following RVT~\cite{goyal2023rvt},
% we utilize a curated subset of $10$ challenging language-conditioned manipulation tasks from RLBench, which includes 166 variations in object properties and scene arrangement.
we train \name{} for $100$k steps, using the LAMB optimizer~\cite{you2019large} optimizer,
with an initial learning rate of 5e-4. 
We also adopt a cosine learning rate decay schedule with warm-up in the first 2k steps.
We use the $\text{SE}(3)$~\cite{shridhar2023perceiver, ze2023gnfactor} augmentation, i.e., translation augmentation of 12.5 cm along the $x$, $y$, and $z$ axis and rotation augmentation of $45^{\circ}$ along the $z$ axis, for expert demonstrations training to enhance the generalizability of policies.
For visual observation, we employ RGB-D images with a resolution of $128\times 128$. 
All the compared methods are trained on 8 NVIDIA A100 GPUs with a batch size of $24$.
We use $96$ demonstrations per task for training and $25$ unseen demonstrations for testing.
Due to the randomness of the sampling-based motion planner, we evaluate each model $10$ times for each task and report the average success rate and standard deviation.
% During the training phase, we use 20 demonstrations for each task.
% There are 100 training demonstrations per task, evenly split across task variations, and 25 (unseen) test episodes for each task.  
% Due to the randomness of the sampling-based motion planner, we evaluate each model four times on the same $25$ variations for each task and report the average success rate and standard deviation.

\mypara{SfM details.}
For the Ego4D dataset, we successfully applied Colmap in most cases, as the moving egocentric view provides sufficient multiple views. 
In challenging cases with sparse views or large motion, Colmap may fail; we then use DUSt3R~\cite{dust3r_cvpr24} as a fallback. 
For the fixed-view Open X-Embodiment dataset, Colmap underperforms, and we instead use DUSt3R with good results.

\subsection{Real World Robot Setup}
% We adopt a total of 5 tasks: \textit{stack blocks}, \textit{press sanitizer}, \textit{put marker in mug/bowl}, \textit{put object in drawer}, \textit{put object in shelf}. Each task can be instantiated with different \textit{variations} defined by the language description. For example, for \textit{stack blocks}, some variations could be ``put yellow block on blue block'' and ``put blue block on red block''.
\mypara{Tasks Setup.}
Our real-world experimental setup consists of a Kinova Gen3 ultra-lightweight robotic arm with two Realsense D435i cameras: one mounted on the wrist to provide a first-person perspective, and the other positioned opposite the robotic arm to offer a third-person view.
We collect RGB-D frame-action data for three tasks,
% In our real-world experiments, we utilized a Kinova Gen3 ultra-lightweight robotic arm with to collect RGB-D frame-action data for six tasks.
% We also employ two Realsense D435i cameras: one mounted on the wrist to provide a first-person perspective, and the other positioned opposite the robotic arm to offer a third-person view.
% We set up both a third-person camera and a wrist camera.
including: ``Pick up”-\textit{Pick up the yellow block}, ``Place in”-\textit{place the yellow block between the blue and red blocks}, and ``Put on”-\textit{put the yellow block to the green subregion}. For each task, we recorded $15$ demonstrations, capturing visual, state, and action data. 
The models are tested with seen/unseen objects to further evaluate generalization capability.

\mypara{Implementation Details.}
We fine-tuned our model on the collected dataset using RDT~\cite{liu2024rdt}, enabling effective generalization across different task variations.
RDT-1B, the largest imitation learning Diffusion Transformer to date, features 1 billion parameters and is pre-trained on over 1 million multi-robot episodes. 
It processes language instructions and RGB images from up to three viewpoints to predict the next 64 robot actions. 
RDT-1B is highly versatile, supporting a wide range of modern mobile manipulators, including single-arm and dual-arm systems, joint-based and end-effector control, position and velocity control, and even wheeled locomotion.
For our task, we primarily adjust the RDT chunk size to 4 and fill its actions to the right-arm portion of the unified action vector, aligning with the RDT pre-training datasets.
Additionally, we set the control frequency of our data to 10. We use an observation window of two frames, which are fed into the head and right wrist image inputs of RDT. The training batch size is set to $20$, while all other settings remain consistent with the official RDT configuration, including a learning rate of 1e-4, the AdamW optimizer, and acceleration via DeepSpeed. We train for $2000$ steps.

\section{Additional Results}

\begin{table}[t]
  \centering \footnotesize
  % \small
  % \vspace{-3pt}
   \addtolength{\tabcolsep}{-2.8pt}
    \begin{tabularx}{\linewidth}{l|c|cccc}
\toprule

 \multirow{2}{*}{\textbf{Method}} &  \textbf{Reconstruction} & \multicolumn{4}{c}{\textbf{Simulation} {\scalebox{0.92}[0.92]{(I2I-CLIP)}}} \\
 \cmidrule(r){3-6} & {\scalebox{0.92}[0.92]{(PSNR/LPIPS/SSIM)}} & \textbf{Remove} & \textbf{Insert} & \textbf{Texture} & \textbf{Color} \\
\midrule

IP2P~\cite{brooks2022instructpix2pix} &  -- & 77.6 & 82.1 & 66.7 & 69.3  \\
\rowcolor{mygray} \bf  \name{}& \bf 40.6 / 0.08 / 0.96 & \bf 93.6 & \bf 97.2 & \bf 92.5 & \bf 92.9  \\
\bottomrule
\end{tabularx}
  \caption{\textbf{Visual metrics.} Avg. of RLBench, Open-X, and Ego4D.}
\label{tab: fid}
\end{table}

\subsection{More Results on Visual Metrics} 
In \cref{tab: fid}, we add reconstruction metric results and, following your advice, include CLIP scores for novel view simulations. Our model consistently outperforms IP2P in visual metrics (see visualizations in \cref{fig:cons}).

\begin{figure}[ht]
    \centering
    \includegraphics[width=1\linewidth]{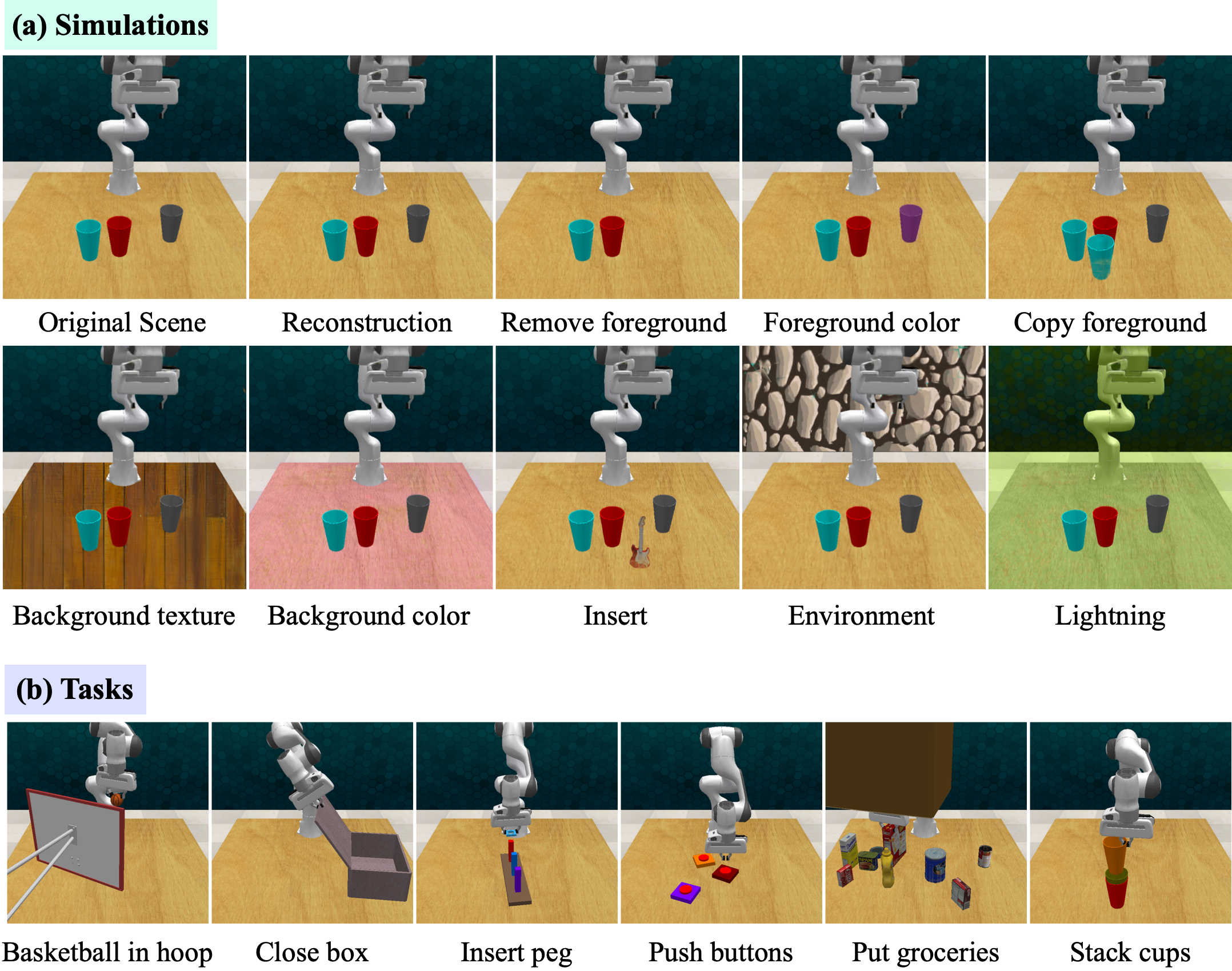}
    %\vspace{-15pt}
    \caption{\textbf{The detailed simulations (a) and more tasks (b) on RLBench} (zoom-in for the best of views). }
    \label{fig:res_supp}
    %\vspace{-10pt}
\end{figure}

\subsection{More Simulation Results}
In \cref{fig:res_supp} (a), we present detailed simulations on the RLBench.
Our \name{} supports a comprehensive set of simulation operators.

\subsection{More Manipulation Tasks Results}
In \cref{fig:res_supp} (b), we show more task demonstrations on the RLBench.
RoboPearls successfully performs multiple manipulation tasks.

\section{Additional Visualizations}
\begin{figure}[t]
    \centering
    \includegraphics[width=1\linewidth]{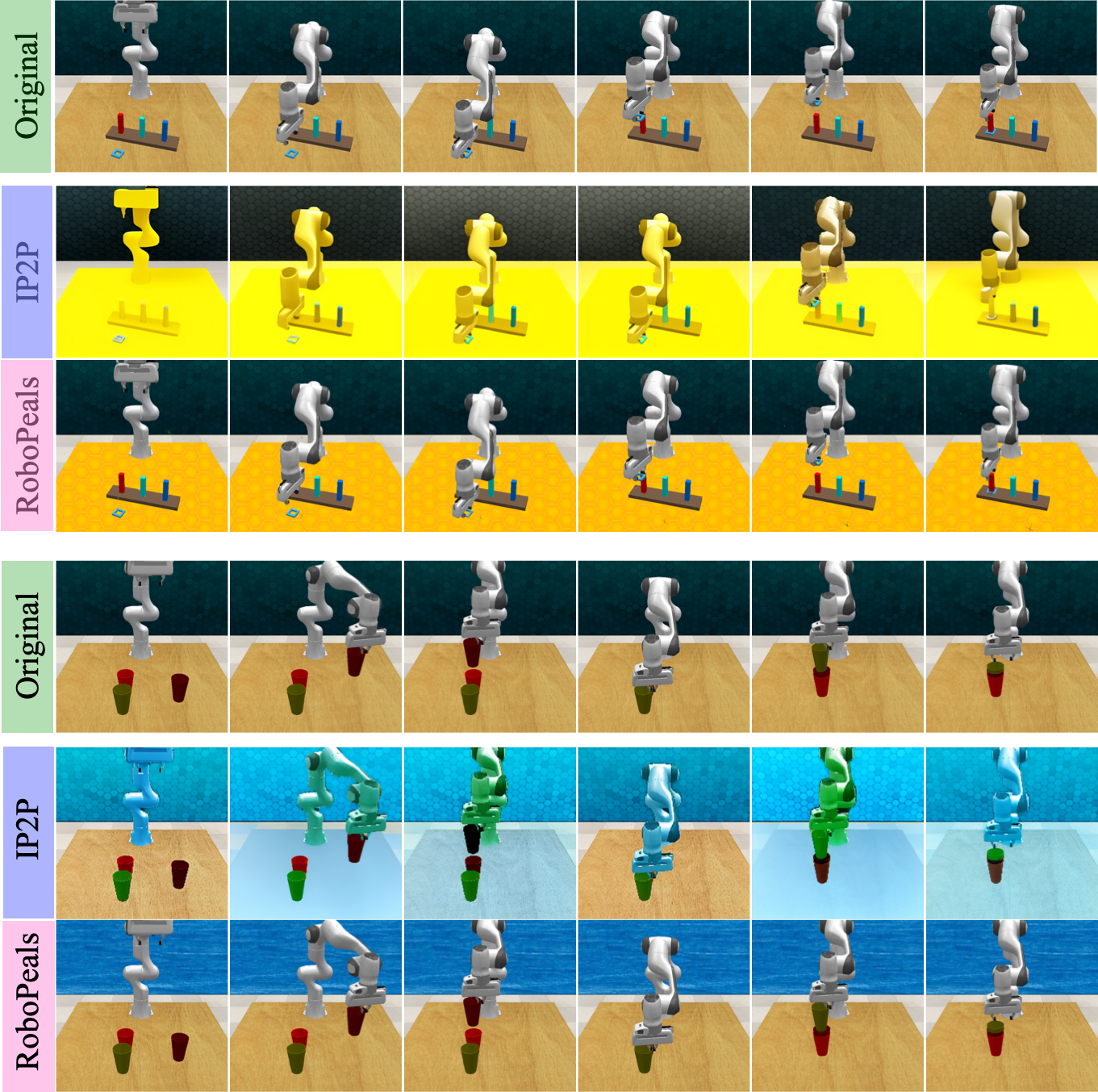}
    % \vspace{-15pt}
    \caption{\textbf{The visualizations of the spatial-temporal consistency} (zoom-in for the best of views). 
}
    \label{fig:cons}
    % \vspace{-10pt}
\end{figure}
\subsection{Spatial-temporal Consistency}
\cref{fig:cons} demonstrates that our simulation maintains significant spatial-temporal consistency, whereas the baseline IP2P struggles.

\begin{figure}[t]
    \centering
    \includegraphics[width=1\linewidth]{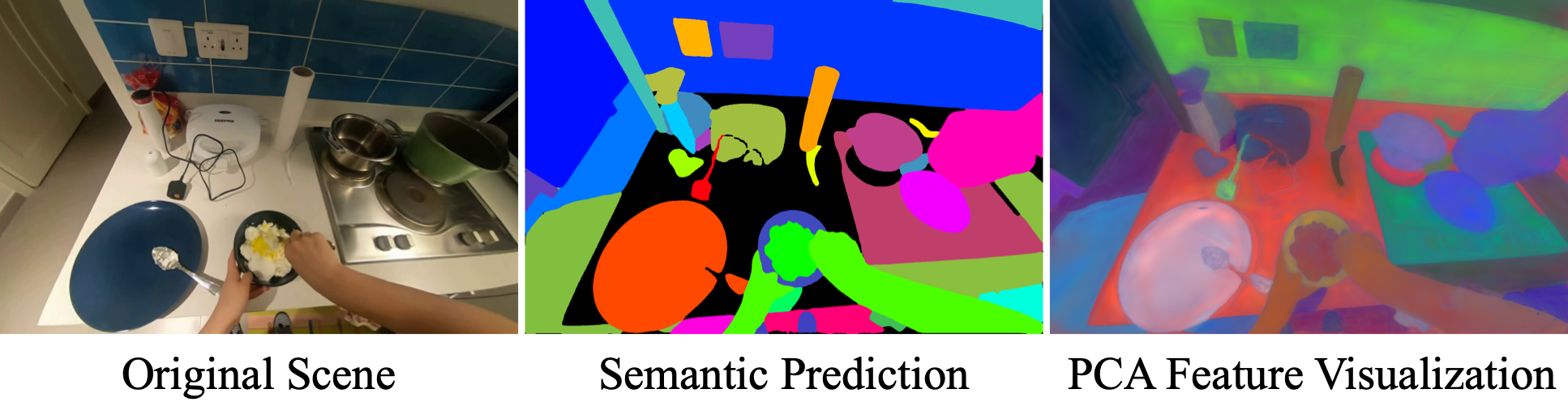}
    % \vspace{-15pt}
    \caption{\textbf{The visualizations of the learned feature vectors.} 
}
    \label{fig:feature}
    % \vspace{-10pt}
\end{figure}
\subsection{Identity Encoding Features}
In \cref{fig:feature}, we adopt PCA to visualize the Identity Encoding features with the rendered semantic and can observe that the approach provides an effective way to select 3D objects in the scene.

% \subsection{Video Demo}
% In addition to the figures, we have attached several video demos in the \href{https://anonymous-github-8ab1cv.github.io/RoboPearls/}{Project Page} and the supplementary material, which consist of hundreds of frames that provide a more comprehensive evaluation of our proposed approach. 
% % The videos also can be found on our
% % \href{https://anonymous-github-8ab1cv.github.io/RoboPearls/}{Project Page}.

% \subsection{Project Page}

% To adhere to the double-blind review process and deadline integrity, our project page is hosted on an \textit{anonymous GitHub Pages} site to ensure the following conditions:
% 1) The image and video materials are too large to include within the supplementary file size limit.
% 2) The hosting site and the linked materials do not reveal the identity or affiliation of the authors.
% 3) The hosting site does not track or identify viewers of the materials.
% 4) We provide a smaller-sized version of our video material in the submitted supplementary material (images in the site are already present in the paper), to ensure that reviewers have a direct way of viewing the material and are also able to verify that the externally hosted material has not been modified since the supplementary material deadline.

% \input{sec_iccv/0_abstract}    
% \input{sec_iccv/1_intro}
% \input{sec_iccv/2_formatting}
% \input{sec_iccv/3_finalcopy}
% {
%     \small
%     \bibliographystyle{ieeenat_fullname}
%     \bibliography{main}
% }

\end{document}